%% file: sample-acmlarge.tex
\documentclass[format=acmsmall, review=false, screen=true]{acmart}
\usepackage{booktabs} 

\usepackage[ruled]{algorithm2e} 

\SetAlFnt{\small}
\SetAlCapFnt{\small}
\SetAlCapNameFnt{\small}
\SetAlCapHSkip{0pt}
\IncMargin{-\parindent}

\usepackage{tikz}
\usepackage{array}
\usepackage{subfig}
\usepackage{xcolor}
\usepackage{amsmath}
\usepackage{mathtools}
\usepackage{verbatim}
\usepackage{url}

\usetikzlibrary{matrix,shadings}
\newcolumntype{V}{>{\centering\arraybackslash}p{6cm}}
\newcommand*{\argmin}{\operatornamewithlimits{arg \min}\limits}
\newcommand*{\argmax}{\operatornamewithlimits{arg \max}\limits}
\def\r{0.1}

\tikzset{
	table/.style={
		matrix of nodes,
		row sep=-\pgflinewidth,
		column sep=-\pgflinewidth,
		nodes={
			rectangle,
			draw=black,
			align=center,
		},
		minimum height=1.5em,
		text depth=0.5em,
		text height=1em,
		text centered,
		nodes in empty cells,
		row 1/.style={
			nodes={
				fill=black,
				text=white,
			}
		},
		rows/.style={nodes={fill=gray!10}},
		columns/.style={nodes={text width = 10em}},
	}
}

\acmJournal{CSUR}
\acmVolume{x}
\acmNumber{x}
\acmArticle{x}
\acmYear{x}
\acmMonth{0}
\acmArticleSeq{x}

\setcopyright{acmcopyright}

\begin{document}
\title{Adversarial Attacks and Defences: A Survey}

\author{Anirban Chakraborty}
\authornote{Corresponding Author}
\affiliation{%
  \institution{Indian Institute of Technology, Kharagpur}
  \streetaddress{Department of Computer Science and Engineering}
  \city{Kharagpur}
  \state{West Bengal}
  \postcode{721302}
  \country{India}}
\email{anirban.chakraborty@iitkgp.ac.in}

\author{Manaar Alam}
\orcid{0000-0002-3338-2944}
\affiliation{%
  \institution{Indian Institute of Technology, Kharagpur}
  \streetaddress{Department of Computer Science and Engineering}
  \city{Kharagpur}
  \state{West Bengal}
  \postcode{721302}
  \country{India}}
\email{alam.manaar@iitkgp.ac.in}

\author{Vishal Dey}
\affiliation{%
  \institution{The Ohio State University, Columbus}
  \streetaddress{Department of Computer Science and Engineering}
  \city{Columbus}
  \state{Ohio}
  \postcode{43210}
  \country{United States}}
\email{dey.78@osu.edu}

\author{Anupam Chattopadhyay}
\affiliation{%
  \institution{Nanyang Technological University}
  \streetaddress{School of Computer Science and Engineering}
  \city{Singapore}
  \postcode{639798}
  \country{Singapore}}
\email{anupam@ntu.edu.sg}

\author{Debdeep Mukhopadhyay}
\affiliation{%
  \institution{Indian Institute of Technology, Kharagpur}
  \streetaddress{Department of Computer Science and Engineering}
  \city{Kharagpur}
  \state{West Bengal}
  \postcode{721302}
  \country{India}}
\email{debdeep@iitkgp.ac.in}

\begin{abstract}
Deep learning has emerged as a strong and efficient framework that can be applied to a broad spectrum of complex learning problems which were difficult to solve using the traditional machine learning techniques in the past. In the last few years, deep learning has advanced radically in such a way that it can surpass human-level performance on a number of tasks. As a consequence, deep learning is being extensively used in most of the recent day-to-day applications. However, security of deep learning systems are vulnerable to crafted adversarial examples, which may be imperceptible to the human eye, but can lead the model to misclassify the output. In recent times, different types of adversaries based on their threat model leverage these vulnerabilities to compromise a deep learning system where adversaries have high incentives. Hence, it is extremely important to provide robustness to deep learning algorithms against these adversaries. However, there are only a few strong countermeasures which can be used in all types of attack scenarios to design a robust deep learning system. In this paper, we attempt to provide a detailed discussion on different types of adversarial attacks with various threat models and also elaborate the efficiency and challenges of recent countermeasures against them.
\end{abstract}

%
%

\ccsdesc[500]{Security and privacy~Software and application security}
\ccsdesc[300]{Computing methodologies~Computer vision}
\ccsdesc[300]{Computing methodologies~Machine learning}

%
%

\keywords{}


\maketitle

\renewcommand{\shortauthors}{A. Chakraborty, M. Alam, V. Dey, A. Chattopadhyay, D. Mukhopadhyay}

\input{samplebody-journals}

\end{document}

%% file: samplebody-journals.tex
\section{Introduction}
    {
	Deep learning is a branch of machine learning that enables computational models composed of multiple processing layers with high level of abstraction to learn from experience and perceive the world in terms of hierarchy of concepts. It uses backpropagation algorithm to discover intricate details in large datasets in order to compute the representation of data in each layer from the representation in the previous layer \cite{lecun2015deep}. Deep learning has been found to be remarkable in providing solutions to the problems which were not possible using conventional machine learning techniques. With the evolution of deep neural network models and availability of high performance hardware to train complex models, deep learning made a remarkable progress in the traditional fields of image classification, speech recognition, language translation along with more advanced areas like analysing potential of drug molecules \cite{ma2015structure}, reconstruction of brain circuits \cite{helmstaedter2013retina}, analysing particle accelerator data \cite{cio2012structure} \cite{kaggle2012higgs}, effects of mutations in DNA \cite{xiong2015gene}. Deep learning network, with their unparalleled accuracy, have brought in major revolution in AI based services on the Internet, including cloud computing based AI services from commercial players like Google \cite{google_cloud}, Alibaba \cite{alibaba_cloud} and corresponding platform propositions from Intel \cite{intel_cloud} and Nvidia \cite{nvidia_cloud}. Extensive use of deep learning based applications can be seen in safety and security-critical environments, like, self driving cars, malware detection and drones and robotics. With recent advancements in face-recognition systems, ATMs and mobile phones are using biometric authentication as a security feature; Automatic Speech Recognition (ASR) models and Voice Controllable systems (VCS) made it possible to realise products like Apple Siri \cite{ios}, Amazon Alexa \cite{alexa} and Microsoft Cortana \cite{cortana}. 
	
    As deep neural networks have found their way from labs to real world, security and integrity of the applications pose great concern. Adversaries can craftily manipulate legitimate inputs, which may be imperceptible to human eye, but can force a trained model to produce incorrect outputs. Szegedy et al. \cite{szegedy2013intriguing} first discovered that well-performing deep neural networks are susceptible to adversarial attacks. Speculative explanations suggested it was due to extreme nonlinearity of deep neural networks, combined with insufficient model averaging and insufficient regularization of the purely supervised learning problem. Carlini et al \cite{carlini2016asr} and Zhang et al \cite{zhang2017vcs} independently brought forward the vulnerabilities of automatic speech recognition and voice controllable systems. Attacks on autonomous vehicles have been demonstrated by Kurakin et al \cite{kurakin2016adversarial} where the adversary manipulated traffic signs to confuse the learning model. The paper by Goodfellow et al. \cite{goodfellow2014explaining} provides a detailed analysis with supportive experiments of adversarial training of linear models, while Papernot et al. \cite{papernot2016transferability} addressed the aspect of generalization of adversarial examples. Abadi et al. \cite{abadi2016deep} introduced the concept of distributed deep learning as a way to protect the privacy of training data. Recently in 2017, Hitaj et al. \cite{hitaj2017deep} exploited the real-time nature of the learning models to train a Generative Adversarial Network and showed that the privacy of the collaborative systems can be jeopardised. Since the findings of Szegedy, a lot of attention has been drawn to the context of adversarial learning and the security of deep neural networks. A number of countermeasures have been proposed in recent years to mitigate the effects of adversarial attacks. Kurakin et al. \cite{kurakin2016adversarial} came up with the idea of using adversarial training to protect the learner by augmenting the training set using both original and perturbed data. Hinton et al. \cite{hinton2015distillating} introduced the concept of distillation which was used by Papernot et al. \cite{papernot2016distillation} to propose a defensive mechanism against adversarial examples. Samangouei et al. \cite{samangouei2018defensegan} proposed a mechanism to use Generative Adversarial Network as a countermeasure for adversarial perturbations. Although each of these proposed defense mechanisms were found to be efficient against particular classes of attacks, none of them could be used as a one-stop solution for all kinds of attacks. Moreover, implementation of these defense strategies can lead to degradation of performance and efficiency of the concerned model. 
    }
	
	
	\subsection{Motivation and Contribution}
	
The importance of Deep learning applications is increasing day-by-day in our daily life. However, these deep learning applications are vulnerable to adversarial attacks. To the best of our knowledge, there has been a little exhaustive survey in the field of adversarial learning covering different types of adversarial attacks and their countermeasures. Akhtar et al.~\cite{akhtar2018threat} presented a comprehensive survey on adversarial attacks on deep learning but in a restrictive context of computer vision. There have been a handful of surveys on security evaluation related to particular machine learning applications~\cite{barreno2006can},~\cite{barreno2010security},~\cite{corona2013adversarial},~\cite{biggio2014securityevaluation}. Kumar et al.~\cite{DBLP:journals/corr/KumarM17} provided a comprehensive survey of prior works by categorizing the attacks under four overlapping classes. The primary motivation of this paper is to summarize recent advances in different types of adversarial attacks with their countermeasures by analyzing various threat models and attack scenarios. We follow a similar approach like prior surveys but without restricting ourselves to specific applications and also in a more elaborate manner with practical examples.
	
	\subsection*{Organization}
	 In this paper, we review recent findings on adversarial attacks and present a detailed understanding of the attack models and methodologies. While our major focus is on attacks and defenses on deep neural networks, we have also presented attack scenarios on Support Vector Machines (SVM) keeping in mind their extensive use in real-world applications. In Section~\ref{sec:taxonomy}, we provide a taxonomy of the related terms and keywords and categorize the threat models. This section also explains adversarial capabilities and illustrates potential attack strategies in training (e.g. poisoning attack) and testing (e.g. evasion attack) phases. We discuss in brief the basic notion of black box and white box attacks with relevant applications and further classify black box attack based on how much information is available to the adversary about the system. Section \ref{sec:exploratory} summarizes exploratory attacks that aim to learn algorithms and models of machine learning systems under attack. Since the attack strategies in evasion and poisoning attacks often overlap, we have combined the work focusing on both of them in Section \ref{sec:evasion_poisoning}. In Section \ref{sec:advancements} we discuss some of the current defense strategies and we conclude in Section \ref{sec:conclusion}. 
	
	\section{Taxonomy of Machine Learning and Adversarial Model}\label{sec:taxonomy}
	Before discussing in details about the attack models and their countermeasures, in this section we will provide a qualitative taxonomy on different terms and key words related to adversarial attacks and categorize the threat models.
	\subsection{Keywords and Definitions}
	In this section, we summarize predominantly used approaches with emphasis on neural networks to solve machine learning problems and their respective application.
	
	\begin{itemize}
	    \item \textbf{Support Vector Machines} Support vector machines (SVMs) are supervised learning models capable of constructing a hyperplane or a set of hyperplanes in high-dimensional space, which can be used for classification, regression or outliers detection. In other words, a SVM model is a representation of data as points in space with objective of building a maximum-margin hyperplane and splitting the training examples into classes, while maximizing the distance between the split points. 
		\item 
		\textbf{Neural Networks:} Artificial Neural networks (ANNs) inspired by the biological neural networks is based on a collection of perceptrons called \emph{neurons}. Each neuron maps a set of inputs to output using an activation function. The learning governs the weights and activation function so as to be able to correctly determine the output. Weights in a multi-layered feed forward are updated by the back-propagation algorithm. Neuron was first introduced by McCulloch-Pitts, followed by Hebb's learning rule, eventually giving rise to multi-layer feed-forward perceptron and backpropagation algorithm. ANNs deal with supervised (CNN, DNN) and unsupervised network models (self organizing maps) and their learning rules. The neural network models used ubiquitously are discussed below.
		
		\begin{enumerate}
			\item 
			\textbf{DNN:} While single layer neural net or perceptron is a feature-engineering approach, deep neural network (DNN) enables feature learning using raw data as input. Multiple hidden layers and its interconnections extract the features from unprocessed input and thus enhances the performance by finding latent structures in unlabeled, unstructured data. A typical DNN architecture, graphically depicted in Figure.~\ref{fig:dnn}, consists of multiple successive layers (at least 2 hidden layers) of neurons. Each processing layer can be viewed as learning a different, more abstract representation of the original multidimensional input distribution. As a whole, a DNN can be viewed as a highly complex function that is capable of nonlinearly mapping original high-dimensional data points to a lower dimensional space.  
			
			\begin{figure}[!h]
				\centering
				\includegraphics[width=0.7\columnwidth]{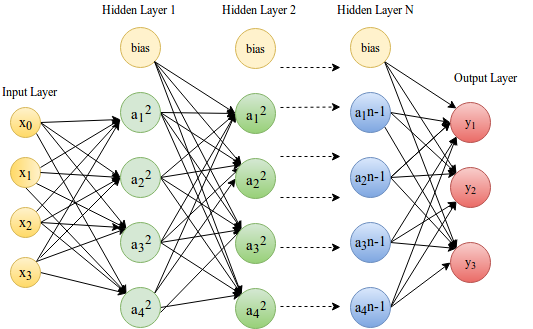}
				\caption{Deep Neural Network}
				\label{fig:dnn}
			\end{figure}
			
			\item
			\textbf{CNN:} A Convolutional Neural Network (CNN) consists of one or more convolutional or sub-sampling layers, followed by one or more fully connected layers, to share weights and reduce the number of parameters. The architecture of CNN, shown in Figure.~\ref{fig:cnn}, is designed to take advantage of 2D input structure (e.g. input image). Convolution layer creates a feature map; pooling (also called sub-sampling or down-sampling) reduces the dimensionality of each feature map but retains the most important informations to have a model robust to small distortions. For example, to describe a large image, feature values in original matrix can be aggregated at various locations (e.g. max-pooling) to form a matrix of lower dimension. The last fully connected layer use the feature matrix formed from previous layers to classify the data. CNN is mainly used for feature extraction, thus it also finds application in data preprocessing commonly used in image recognition tasks.
			\begin{figure}[!h]
				\centering
				\includegraphics[width=\linewidth, height=6cm]{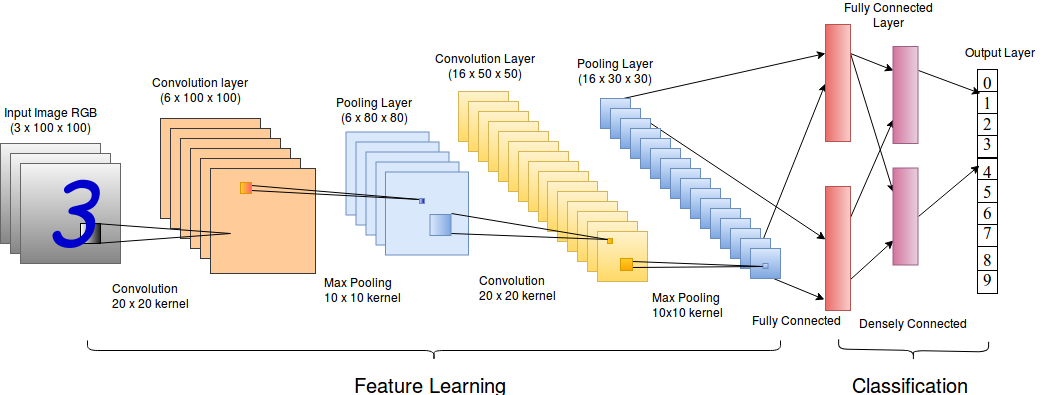}
				\caption{Convolutional Neural Network for MNIST digit recognition}
				\label{fig:cnn}
			\end{figure}
		\end{enumerate}
	\end{itemize}
	
	\subsection{Adversarial Threat Model}\label{ssec:threat_model}
	The security of any machine learning model is measured concerning the adversarial goals and capabilities. In this section, we taxonomize the threat models in machine learning systems keeping in mind the strength of the adversary. We begin with the identification of threat surface~\cite{papernot2016towards} of systems built on machine learning models to identify where and how an adversary may attempt to subvert the system under attack.
	
	\subsubsection{The Attack Surface}
\begin{figure}[!h]
		\centering
		\includegraphics[width=\textwidth]{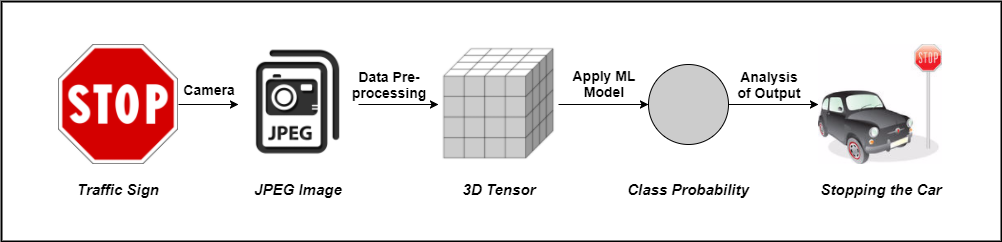}
		\caption{Generic pipeline of an Automated Vehicle System\label{fig:automated_vehicle}}
	\end{figure}

	A system built on Machine Learning can be viewed as a generalized data processing pipeline. A primitive sequence of operations of the system at the testing time can be viewed as: a) collection of input data from sensors or data repositories, b) transferring the data in the digital domain, c) processing of the transformed data by machine learning model to produce an output, and finally, d) action taken based on the output. For illustration, consider a generic pipeline of an automated vehicle system as shown Figure~\ref{fig:automated_vehicle}.
	
	The system collects sensor inputs (images using camera) from which model features (tensor of pixel values) are extracted and used within the models. It then interprets the meaning of the output (probability of stop sign), and takes appropriate action (stopping the car). The \emph{attack surface}, in this case, can be defined with respect to the data processing pipeline. An adversary can attempt to manipulate either the \emph{collection} or the \emph{processing} of data to corrupt the target model, thus tampering the original output. The main attack scenarios identified by the attack surface are sketched below~\cite{biggio2014security, biggio2014securityevaluation}:
	\begin{enumerate}
		\item \textit{Evasion Attack}: This is the most common type of attack in the adversarial setting. The adversary tries to evade the system by adjusting malicious samples during testing phase. This setting does not assume any influence over the training data. 
		
		\item \textit{Poisoning Attack}: This type of attack, known as contamination of the training data, takes place during the training time of the machine learning model. An adversary tries to poison the training data by injecting carefully designed samples to compromise the whole learning process eventually.
		
		\item \textit{Exploratory Attack}: These attacks do not influence training dataset. Given black box access to the model, they try to gain as much knowledge as possible about the learning algorithm of the underlying system and pattern in training data. 
	\end{enumerate}
	
	The definition of a threat model depends on the information the adversary has at their disposal. Next, we discuss in details the adversarial capabilities for the threat model.
	
	\subsubsection{The Adversarial Capabilities}
	\label{sssec: adversarial_capa}
	The term adversarial capabilities refer to the amount of information available to an adversary about the system, which also indicates the attack vector he may use on the threat surface. For illustration, again consider the case of an automated vehicle system as shown in Figure~\ref{fig:automated_vehicle} with the attack surface being the testing time (i.e., an Evasion Attack). An internal adversary is one who have access to the model architecture and can use it to distinguish between different images and traffic signs, whereas a weaker adversary is one who have access only to the dump of images fed to the model during testing time. Though both the adversaries are working on the same attack surface, the former adversary is assumed to have much more information and is thus strictly ``stronger''. We explore the range of adversarial capabilities in machine learning systems as they relate to testing and training phases.
	
	\subsubsection*{Training Phase Capabilities}
	Attacks during training time attempt to influence or corrupt the model directly by altering the dataset used for training. The most straightforward and arguably the weakest attack on training phase is by merely accessing a partial or full training data. There are three broad attack strategies for altering the model based on the adversarial capabilities.
	
	\begin{enumerate}
		\item \textit{\textbf{Data Injection:}} The adversary does not have any access to the training data as well as to the learning algorithm but has ability to augment a new data to the training set. He can corrupt the target model by inserting adversarial samples into the training dataset.  
		
		\item \textit{\textbf{Data Modification:}} The adversary does not have access to the learning algorithm but has full access to the training data. He poisons the training data directly by modifying the data before it is used for training the target model.
		
		\item \textit{\textbf{Logic Corruption:}} The adversary has the ability to meddle with the learning algorithm. These attacks are referred as logic corruption. Apparently, it becomes very difficult to design counter strategy against these adversaries who can alter the learning logic, thereby controlling the model itself.
	\end{enumerate}
	
	\subsubsection*{Testing Phase Capabilities}
	\label{sssec:testing_capb}
	Adversarial attacks at the testing time do not tamper with the targeted model but rather forces it to produce incorrect outputs. The effectiveness of such attacks is determined mainly by the amount of information available to the adversary about the model. Testing phase attacks can be broadly classified into either \emph{White-Box} or \emph{Black-Box} attacks. Before discussing these attacks, we provide a formal definition of a training procedure for a machine learning model.
	
	Let us consider a target machine learning model $f$ is trained over input pair $(X, y)$ from the data distribution $\mu$ with a randomized training procedure $train$ having randomness $r$ (e.g., random weight initialization, dropout, etc.). The model parameters $\theta$ are learned after the training procedure. More formally, we can write:
	
	\begin{equation*}
	\centering
	\theta \leftarrow train(f, X, y, r)
	\end{equation*}
	
	Now, let us understand the capabilities of the white-box and black-box adversaries with respect to this definition. An overview of the different threat models have been shown in Figure.~\ref{tab:table2}
	
	\begin{figure}[!h]
		\centering
		\begin{tikzpicture}
		\matrix[table, rows={2,...,10}{fill=grey!10}, columns={1,...,4}{text width = 8em}, ampersand replacement=\&] (first)
		{
			Article \& Black box\& White box\\
			Papernot, Nicolas\cite{papernot2016transferability} \& Non-adaptive \& \\
			Rosenberg, Ishai \cite{rosenberg2017generic} \& Adaptive \&\\
			Tram\`er, Florian \cite{tramer2016stealing} \& Non-Adaptive \& \\
			Papernot, Nicolas \cite{papernot2017practical} \& (Non)-Adaptive \& \\
			Fredrikson, Matt \cite{fredrikson2015model} \& Adaptive \& \\
			Shokri, Reza \cite{shokri2017membership} \& Adaptive \&\\
			Hitaj, Briland \cite{hitaj2017deep} \& Strict \& \\
			Moosavi-Dezfooli \cite{moosavi2016deepfool} \& \& \\
			Tram\`er, Florian \cite{Tramer2017ensemble} \& \& \\
		};
		
		\fill[black] (first-8-3) circle (\r);
		\fill[black] (first-10-3) circle (\r);
		\end{tikzpicture}
		\caption{Overview of threat models in relevant articles}
		\label{tab:table2}
	\end{figure}
	\subsubsection*{White-Box Attacks}
	In white-box attack on a machine learning model, an adversary has total knowledge about the model ($f$) used for classification (e.g., type of neural network along with number of layers). The attacker has information about the algorithm ($train$) used in training (e.g., gradient-descent optimization) and can access the training data distribution ($\mu$). He also knows the parameters ($\theta$) of the fully trained model architecture. The adversary utilizes available information to identify the feature space where the model may be vulnerable, i.e, for which the model has a high error rate. Then the model is exploited by altering an input using adversarial example crafting method, which we discuss later. The access to internal model weights for a white-box attack corresponds to a very strong adversarial attack.
	
	\subsubsection*{Black-Box Attacks}\label{black-box}
	Black-Box attack, on the contrary, assumes no knowledge about the model and uses information about the settings or past inputs to analyse the vulnerability of the model. For example, in an oracle attack, the adversary exploits a model by providing a series of carefully crafted inputs and observing outputs. Black Box attacks can be further classified into the following categories:
	\begin{enumerate}
		\item \textit{\textbf{Non-Adaptive Black-Box Attack:}} For a target model ($f$), a non-adaptive black-box adversary only gets access to the target model's training data distribution ($\mu$). The adversary then chooses a procedure $train^{\prime}$ for a model architecture $f^{\prime}$ and trains a local model over samples from the data distribution $\mu$ to approximate the model learned by the target classifier. The adversary crafts adversarial examples on the local model $f^{\prime}$ using white-box attack strategies and applies these crafted inputs to the target model to force mis-classifications.
		\item \textit{\textbf{Adaptive Black-Box Attack:}} For a target model ($f$), an adaptive black-box adversary does not have any information regarding the training process but can access the target model as an oracle. This strategy is analogous to chosen-plaintext attack in cryptography. The adversary issues adaptive oracle queries to the target model and labels a carefully selected dataset, i.e., for any arbitrarily chosen $x$ the adversary obtains its label $y$ by querying the target model $f$. The adversary then chooses a procedure $train^{\prime}$ and model architecture $f^{\prime}$ to train a surrogate model over tuples $(x, y)$ obtained from querying the target model. The surrogate model then produces adversarial samples by following white-box attack technique for forcing the target model to mis-classify malicious data.
		\item \textit{\textbf{Strict Black-Box Attack:}} A black-box adversary sometimes may not contain the data distribution $\mu$ but has the ability to collect the input-output pairs $(x, y)$ from the target classifier. However, he can not change the inputs to observe the changes in output like an adaptive attack procedure. This strategy is analogous to the known-plaintext attack in cryptography and would most likely to be successful for a large set of input-output pairs.
	\end{enumerate}
	
	The point to be remembered in the context of a black-box attack is that an adversary neither tries to learn the randomness $r$ used to train the target model nor the target model's parameters $\theta$. The primary objective of a black-box adversary is to train a local model with the data distribution $\mu$ in case of a non-adaptive attack and with carefully selected dataset by querying the target model in case of an adaptive attack. Table~\ref{tab:table1} shows a brief distinction between black box and white box attacks.
	\begin{table}[!h]
		\centering
		\begin{tabular}{l|c|V}
			\textbf{Description} & \textbf{Black box attack} & \textbf{White box attack}\\
			\hline \hline
			Adversary & Restricted knowledge from being & Detailed knowledge of the network\\Knowledge & able to only observe the networks & architecture and the parameters\\ &
			output on some probed inputs.&
			resulting from training.\\
			\hline
			Attack
			& Based on a greedy local search & Based on the gradient of the network\\ Strategy & generating an implicit approximation & loss function w.r.t to the input.\\&to the actual gradient w.r.t \\& the current output by observing\\& changes in input.
			
		\end{tabular}
		\caption{Distinction between black box and white box attacks}
		\label{tab:table1}
	\end{table}

	The adversarial threat model also depends not only depends on the adversarial capabilities but also on the action taken by the adversary. In the next subsection, we discuss the goal of an adversary while compromising the security of any machine learning system.
	
	\subsubsection{Adversarial Goals}
	An adversary attempts to provide an input $x_{*}$ to a classification system that results in an incorrect output classification. The objective of the adversary is inferred from the incorrectness of the model. Based on the impact on the classifier output integrity the adversarial goals can be broadly classified as follows:
	
	\begin{enumerate}
		\item \textit{\textbf{Confidence Reduction:}} The adversary tries to reduce the confidence of prediction for the target model. For example, a legitimate image of a `stop' sign can be predicted with a lower confidence having a lesser probability of class belongingness.
		\item \textit{\textbf{Misclassification:}} The adversary tries to alter the output classification of an input example to any class different from the original class. For example, a legitimate image of a `stop' sign will be predicted as any other class different from the class of stop sign.
		\item \textit{\textbf{Targeted Misclassification:}} The adversary tries to produce inputs that force the output of the classification model to be a specific target class. For example, any input image to the classification model will be predicted as a class of images having `go' sign.
		\item \textit{\textbf{Source/Target Misclassification:}} The adversary attempts to force the output of classification for a specific input to be a particular target class. For example, the input image of `stop' sign will be predicted as `go' sign by the classification model.
	\end{enumerate}
	
\begin{figure}[!t]
		\begin{minipage}[b]{0.45\textwidth}
		    \centering
		    \includegraphics[width=\textwidth]{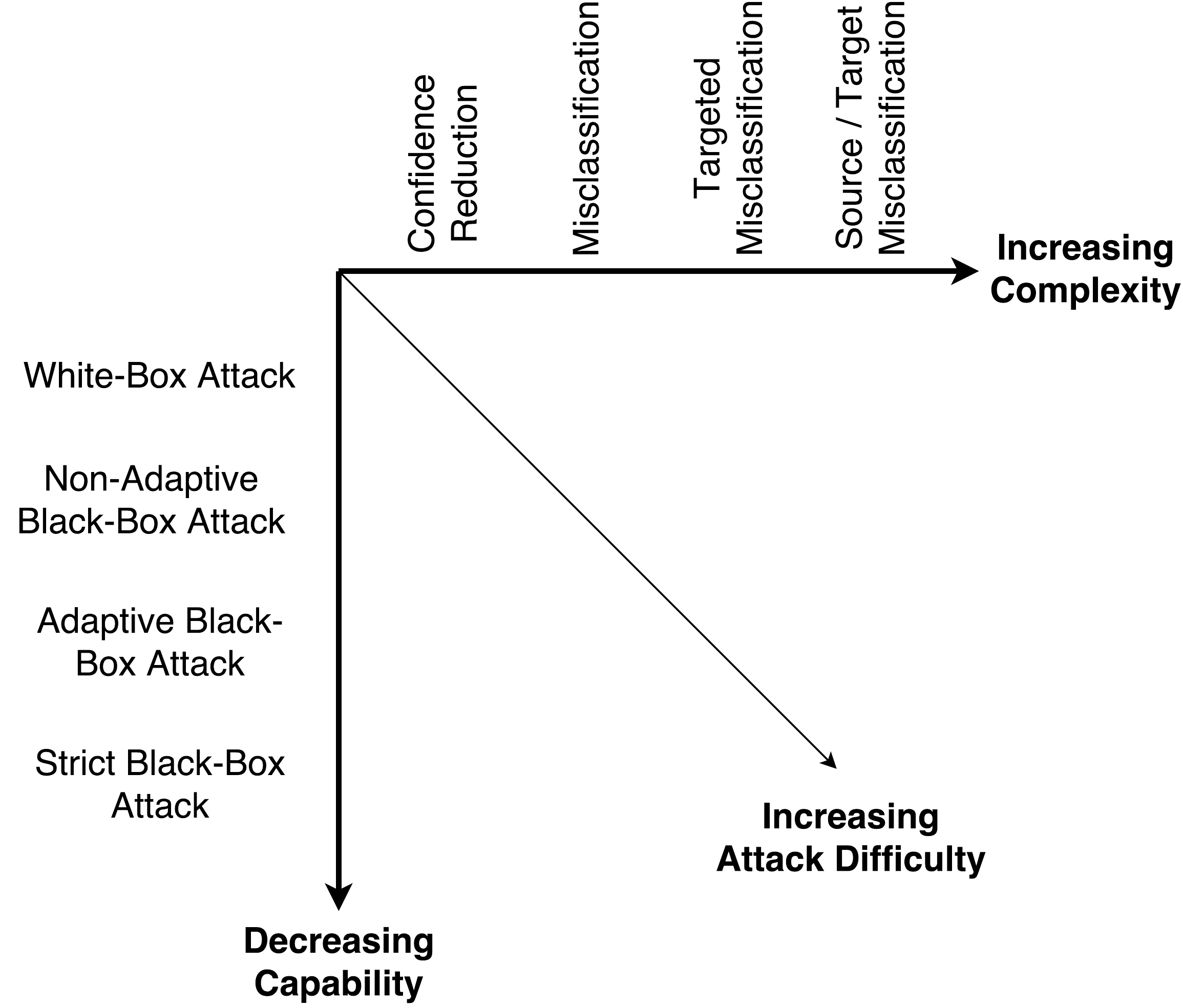}
		    \caption*{a) Attack Difficulty with respect to adversarial capabilities and goals for Evasion Attacks}
		    \label{fig:evasion}
		\end{minipage}
		\qquad
		\begin{minipage}[b]{0.45\textwidth}
		    \centering
		    \includegraphics[width=\textwidth]{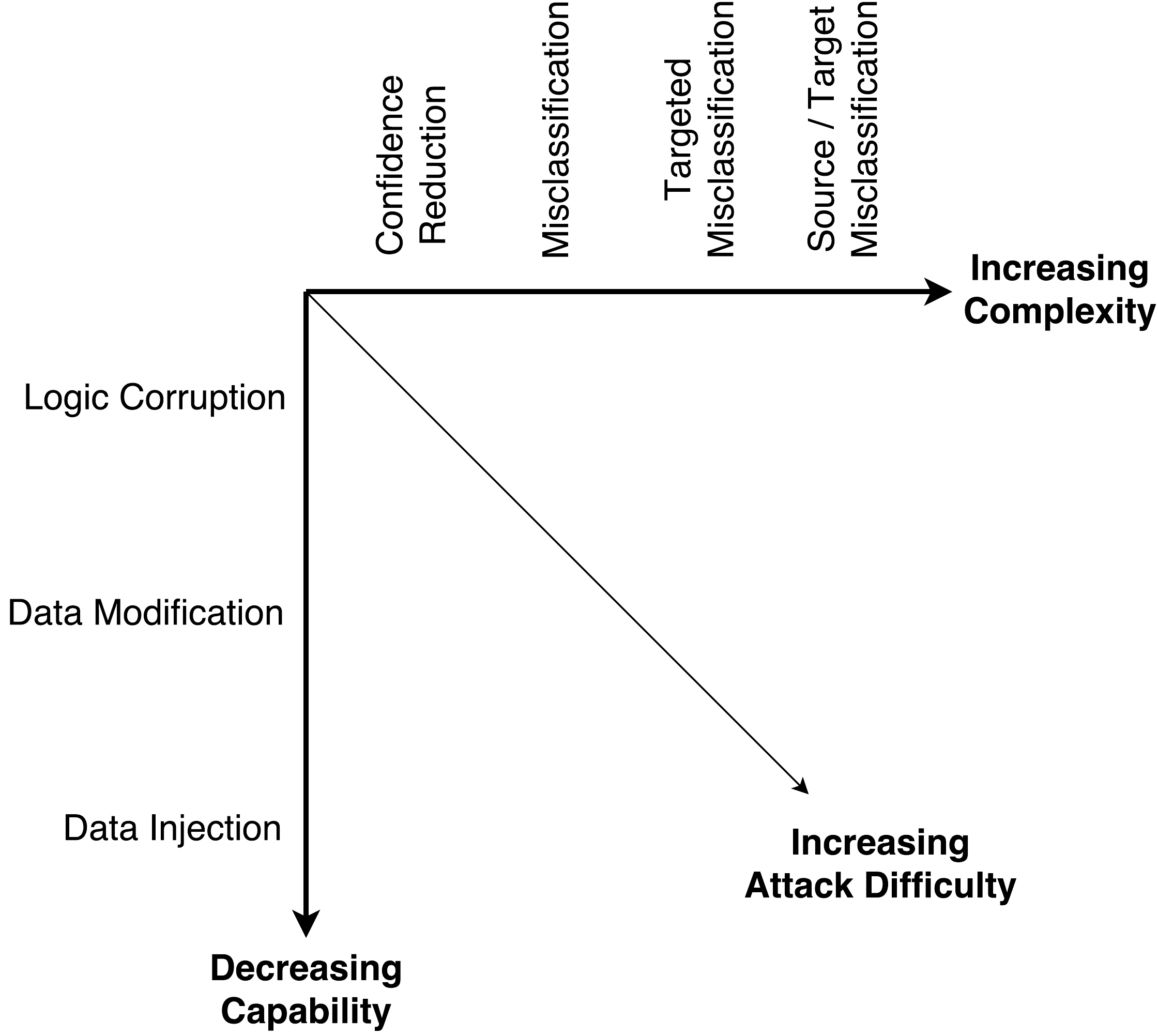}
		    \caption*{b) Attack Difficulty with respect to adversarial capabilities and goals for Poisoning Attacks}
		    \label{fig:poisoning}
		\end{minipage}
	    \caption{Taxonomy of Adversarial Model for a) Evasion Attacks and b) Poisoning Attacks with respect to adversarial capabilities and goals}
	    \label{fig:taxonomoy}
	\end{figure}

	The taxonomy of the adversarial threat model for both the evasion and the poisoning attacks with respect to the adversarial capabilities and adversarial goals are represented graphically in Figure~\ref{fig:taxonomoy}. The horizontal axis of both figures represents the complexity of adversarial goals in increasing order, and the vertical axis loosely represents the strength of an adversary in decreasing order. The diagonal axis represents the complexity of a successful attack based on the adversarial capabilities and goals.
	
	Some of the noteworthy attacks along with their target applications is shown in Table.~\ref{tab:table3}. Further in Table.~\ref{tab:table_summary}, we categorize those attacks under different threat models and discuss in detail about them in the next section.
	
	\begin{table}[!h]
		\centering
		\begin{tabular}{l|c|V}
			\textbf{Articles} & \textbf{Attacks} & \textbf{Applications}\\
			\hline
			Fredrikson et al. \cite{fredrikson2015model} & Model Inversion & Biomedical Imaging, \\ && biometric identification\\
			\hline
			Tram\`er et al. \cite{tramer2016stealing} & Extraction of target machine & Attacks extend to multiclass  \\ & learning models using APIs & classifications \& neural networks\\
			\hline
			Anteniese et al. \cite{AtenieseFMSVV13} & Meta-classifier to hack & Speech Recognition\\ & other classifiers & \\
			\hline
			Biggio et al. \cite{Biggio2011support}, \cite{Biggio2012poisoning}& Poisoning based attacks: & Crafted training data for\\ & & Support vector Machines\\
			\hline
			Dalvi et al. \cite{dalvi2004adversarial} & Adversarial Classification, & Email Spam detection, fraud \\Biggio et al. \cite{biggio2008adversarial}, \cite{biggio2014svm}& Pattern recognition & detection, intrusion detection, \\ & & biometric identification\\
			\hline
			Papernot et al. & Adversarial samples crafting, & digit recognition, black-box \\ \cite{papernot2017practical}, \cite{papernot2016transferability} & adversarial sample & attacks against classifiers hosted\\ &transferability  & by Amazon and Google \\
			\hline
			Hitaj et al. \cite{hitaj2017deep} & GAN under collaborative learning & Classification
			\\
			\hline
			Goodfellow et al. \cite{goodfellow2014gan} & Generative Adversarial Network & Classifiers, Malware Detection\\ 
			\hline
			Shokri et al. \cite{shokri2017membership} & Membership inference attack & Attack on classification models trained\\ & & by
			commercial "ML as a service" providers such as Google and Amazon\\
			\hline
			Moosavi et al. \cite{moosavi2016deepfool} & Adversarial perturbations: & Image classification \\ Carlini et al. \cite{carlini2017towards} & and sample generation: & intrusion detection \\Li et al. \cite{Li2016data} & Poisoning based attack & Collaborative filtering systems
		\end{tabular}
		\caption{Overview of Attacks and Applications}
		\label{tab:table3}
	\end{table}
	
	\begin{table}[!b]
		\centering
		\begin{tabular}{l|p{70mm}}
			\hline
			Exploratory Attacks &  Model Inversion
			\newline Membership Inference attack
			\newline Model Extraction via APIs
			\newline Information Inference
			\\
			\hline
			Evasion Attacks &
			Adversarial Examples Generation
			\newline Generative Adversarial Networks (GAN) 
			\newline GAN based attack in collaborative learning
			\newline Intrusion Detection Systems
			\newline Adversarial Classification
			\\
			\hline
			Poisoning Attacks &
			Support Vector Machine Poisoning
			\newline
			Poisoning on collaborative filtering systems
			\newline
			Anomaly Detection Systems
			\\
			\hline
		\end{tabular}
	    \caption{Attack Summary}
		\label{tab:table_summary}
	\end{table}
	
	\section{Exploratory Attacks}\label{sec:exploratory}
	Exploratory attacks do not modify the training set but instead tries to gain information about the state by probing the learner. The adversarial examples are crafted in such a way that the learner passes them as legitimate examples during testing phase.
	
	\subsection{Model Inversion(MI) Attack}
	Fredrikson et al. introduced "model inversion" in \cite{fredrikson2014privacy} where they considered a linear regression model $f$ that predicted drug dosage using patient information, medical history and genetic markers; explored that given white-box access to model $f$ and an instance of data $(\mathbf{X} = \{x_1, x_2,... ,x_n\}, y)$, model inversion infers genetic marker $x_1$. The algorithm produces least-biased maximum a priori (MAP) estimate for $x_1$ by iterating over all possible values of nominal feature ($x_1$) for obtaining target value $y$, thus minimizing adversary's mis-prediction rate. It has serious limitations; for e.g. it cannot handle larger set of unknown features since it is computationally not feasible. 
	
\begin{figure}[!t]
		\centering
		\includegraphics*[width = 0.7\textwidth]{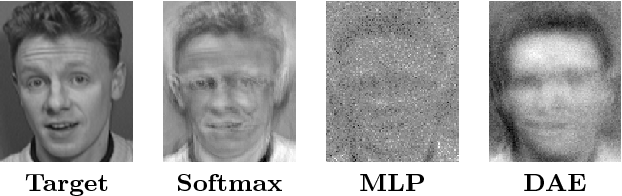}
		\caption{Reconstruction of the individual on the left by Softmax, MLP, and DAE}\vspace*{-3mm}
		\caption*{(Image Credit: Fredrikson et al. \cite{fredrikson2015model})}
		\label{model_inversion}
	\end{figure}

	Fredrikson et al. \cite{fredrikson2015model} intended to remove limitations of their previous work and showed that for a black-box model, an attacker can predict the patient's genetic markers. This new model inversion attacks through ML APIs that exploit confidence values in a variety of settings and explored countermeasures in both black box and white box settings. The attack infers sensitive features used as inputs to decision tree models for lifestyle surveys, as well as to recover images from API access to facial recognition services. This attack has been successfully experimented in face recognition using neural network models: softmax regression, multilayer perceptron (MLP) and stacked denoising autoencoder network (DAE); given access to the model and person's name, it can recover the facial image. The reconstruction produced by the three algorithms is shown in Figure.~\ref{model_inversion}. Due to the rich structure of deep learning machines, the model inversion attack may recover only prototypical examples that have little resemblance to the actual data that defined the class. 
	
	\subsection{Model Extraction using APIs}
	Tram\`er et al. \cite{tramer2016stealing} presented simple attacks to extract target machine learning models for popular model classes such as logistic regression, neural networks, and decision trees. Attacks presented are strict black box attacks, but could build models locally that are functionally close to target.The authors demonstrated Model Extraction attack on online ML service providers such as BigML and Amazon Machine Learning. Machine learning APIs provided by ML-as-service providers return precision confidence values along with class labels. Since the attacker do not have any information regarding the model or training data distribution, he can attempt to solve mathematically for unknown parameters or features given the confidence value and equations by quering $d+1$ random $d$-dimensional inputs for unknown $d+1$ parameters.  
	
	\subsection{Inference Attack}
	Ateniese et al. \cite{AtenieseFMSVV13} showed it is possible to gather relevant information from machine learning classifiers using a meta-classifier. Given the black box access to a model (e.g., via public APIs) and a training data, an attacker may be interested in knowing whether that data was part of the training set of the model. They experimented with a speech recognition classifier that uses Hidden Markov Models and extracted information such as accent of the users which was not supposed to be explicitly captured.
	
	Another inference attack presented by Shokri et al. \cite{shokri2017membership} is membership inference, i.e., which determines whether a given data point belongs to the same distribution as the training dataset. This attack may fall under the category of non-adaptive or adaptive black box attacks. In a typical black box environment, attacker sends a query to the target model with a data point and obtains model's prediction. The output given by the model is a vector of probabilities which specifies whether the data point belongs to a certain class. For training attack model, a set of shadow models are built. Since the adversary has the knowledge of whether a given record belongs to the training set, supervised learning can be employed and corresponding output labels are then fed to attack model to train it to distinguish shadow model's outputs on members of their training data from those of non-members.  
	
	\begin{figure}[!t]
		\centering
		\includegraphics[width=0.8\textwidth]{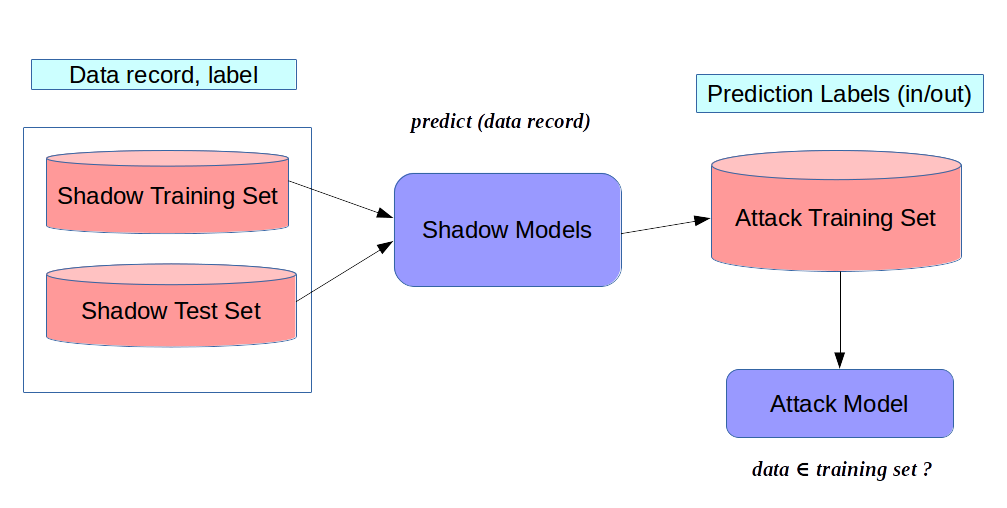}
		\caption{Overview of Membership inference attack in the black-box setting}
		\label{fig: inference}
	\end{figure}
	Figure \ref{fig: inference} illustrates the end-to-end attack process.
	The output vectors obtained from shadow model are labeled ``in'' and added to the attack model's training dataset. A test dataset is also used to query the shadow model and the outputs from this set are labeled ``out'' and also added to the attack model's training dataset. Thus a collection of target attack models is trained by utilizing the black box behaviour of the shadow models.
	Authors used membership attacks on classification models trained by commercial "ML as a service" providers such as Google and Amazon. 
	
	
	\section{Evasion \& Poisoning Attacks}
	\label{sec:evasion_poisoning}
	Evasion attacks are the most common attacks on machine learning systems. Malicious inputs are craftily modified so as to force the model to make a false prediction and evade detection. Poisoning attack differs in that the inputs are modified during training and model is trained on contaminated inputs to obtain desired output.
	
	\subsection{Generative Adversarial Attack(GAN)}
\begin{figure}[!t]
			\centering
			\includegraphics[width=\textwidth]{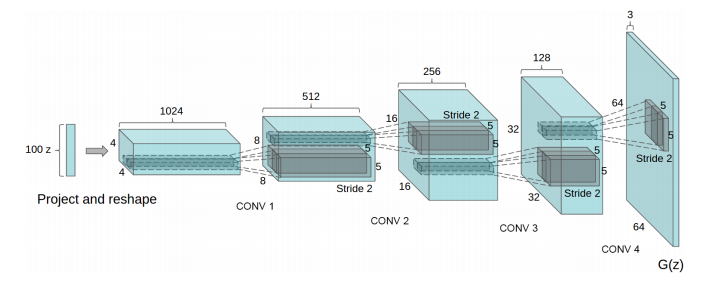}
			\caption{DCGAN generator used for LSUN scene modeling.\label{fig:cdgan}} \vspace*{-3mm}
			\caption*{(Image Credit: Radford et al. \cite{radford2018dcgan})}
		\end{figure}

	{ Goodfellow et al \cite{goodfellow2014gan} introduced generative adversarial networks whose aim is to generate samples similar to the training set, having almost identical distribution. 
	The GAN procedure, as depicted in Figure.~\ref{fig:gan}, is composed of a discriminative deep learning network $D$ and a generative deep learning network $G$. The role of discriminative network is to distinguish between samples taken from the original database and those generated by GAN. The generative network is first initialized with random noise. Its role is to produce samples identical to the training set of the discriminator. Formally, $G$ is trained to maximize the probability of $D$ making a mistake. This competition leads both the models to improve their accuracy. The procedure ends when $D$ fails to distinguish between samples from the training set and those generated by $G$. The entities and adversaries are in a constant duel where one (generator) tries to fool the other(discriminator), while the other tries to prevent being fooled. 
		
		\begin{figure}[!t]
			\centering
			\includegraphics[scale=0.6]{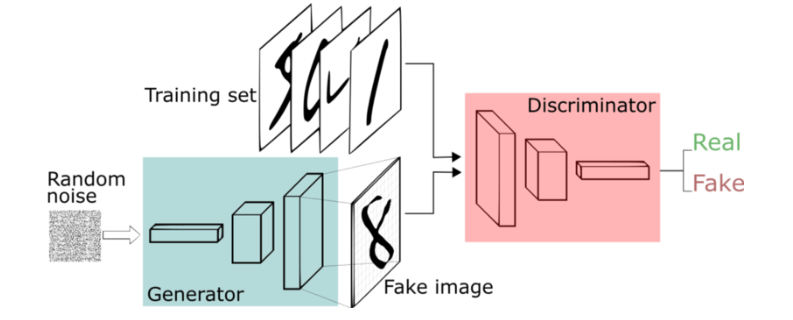}
			\caption{Generative Adversarial Learning\label{fig:gan}} \vspace*{-3mm}
			\caption*{(Image Credit: Thalles Silva \cite{medium})}
			\label{fig:gan}
		\end{figure}
		
	The authors (Goodfellow et al \cite{goodfellow2014gan}) defined the value function \textit{$V(G,D)$} as 
	\[ \min_{G} \max_{D} V(D,G) = \mathbb{E}_{\textit{x}\sim p_{data}{(\textit{x})}} \log D(\textit{x}) + \mathbb{E}_{\textit{z}\sim p_{z}{(\textit{z})}} \log (1 - D(G(\textit{z}))) \]
	where $p_g(x)$ is the generator's distribution, $p_z(z)$ is a prior on input noise variables. The objective is to train $D$ to maximize the probability of assigning correct label to training and sample examples, while simultaneously training $G$ to minimize it. It was found that optimizing $D$ was computationally intensive and on finite data sets, there was a chance of over fitting. Moreover, during the initial stages of learning when $G$ is poor, $D$ can reject samples with relatively high confidence as it can distinguish them from training data. So, instead of training $G$ to minimize $\log {{(1 - D(G(z))}}$, they trained $G$ to maximize $\log {(D(G(z))}$.
	
	Radford et al \cite{radford2018dcgan} introduced a new class of CNNs called Deep Convolutional Generative
    Adversarial Networks (DCGANs) which overcomes these constraints in GANs and make them stable to train. Some of the key insights of DCGAN architecture were:
        \begin{itemize}
            \item The overall network architecture was based on all convolutional net \cite{springenberg2014convnet} replacing deterministic pooling functions with strided convolutions.
            \item The first layer of the GAN, which takes random noise as input, was reshaped into 4-dimensional tensor and used as the start of convolutional stack. The last convolutional layer was flattened and fed into a single sigmoid output (Figure. ~\ref{fig:cdgan}). \cite{mordvintsev2015inception}
            \item They used batch normalization \cite{sergey2015batchnorm}
            to stabilize the learning by normalizing the input in order to have zero mean and unit variance. This solved the problem of instability of GAN during training which arise due to poor initialization.
            \item ReLU activation \cite{nair2010relu} was used in all layers in generator, except for the output layer which uses Tanh function. Leaky ReLU activation \cite{mass2013leakyrelu}, \cite{xu2015leakyrelu} was used for all layers in discriminator.
        \end{itemize}
	
	}

	\subsection{Adversarial Examples Generation} \label{advex_gen}
	In this section, we present a general overview for modifying samples so that a classification model yields an adversarial output. The adversarial sample modification can be performed both in training and testing phase.
	
	\subsubsection{Training Phase Modification}
	A learning process fine-tunes the parameters $\theta$ of the hypothesis $h$ by analyzing a training set. This makes the training set susceptible to manipulation by adversaries. Barreno et al.\cite{barreno2006can} first proposed the term \emph{Poisoning Attacks} which alters the training dataset by inserting, modifying or deleting points keeping the intention of modifying the decision boundaries of the targeted model following the work of Kearns et al.~\cite{Kearns1993learning}, thus challenging the learning system's integrity. The poisoning attack of the training set can be done in two ways: either by direct modification of the labels of the training data or by manipulating the input features depending on the capabilities posed by the adversary. We present a brief overview of both the techniques without much technical details, as the training phase attack needs more powerful adversary and thus is not common in general.
	
	\begin{itemize}
		\item \textit{Label Manipulation}
		If the adversary has the capability to modify the training labels only, then he must obtain the most vulnerable label given the full or partial knowledge of the learning model. A basic approach is to randomly perturb the labels, i.e., select new labels for a subset of training data by picking from a random distribution. Biggio et al.~\cite{Biggio2011support} presented a study which shows that a random flip of 40\% of the training labels is sufficient to degrade the performance of classifiers learned with SVMs.
		
		\item \textit{Input Manipulation}
		In this scenario, the adversary is more powerful and can corrupt the input features of training points analyzed by the learning algorithm, in addition to its labels. This scenario also assumes that the adversary has the knowledge of the learning algorithm.
		
		Kloft et al.~\cite{kloft2010online} presented a study in which they showed that inserting malicious points in the training dataset could gradually shift the decision boundary of an anomaly detection classifier. The learning algorithm that they used for the study works in an online scenario - new training data are collected at regular intervals, and the parameter values $\theta$ are fine-tuned based on a sliding window of that data. Thus, injecting new points in the training dataset is essentially an easy task for the adversary. Poisoning data points can be obtained by solving a linear programming problem which has the objective of maximizing displacement of the mean of the training data.
		
		In the offline learning settings, Biggio et al.~\cite{Biggio2012poisoning} introduced an attack that inserts inputs in the training set, which are crafted using a gradient ascent method. The method identifies the inputs corresponding to local maxima in the test error of the model. They presented a study which shows that by including these inputs into the training set one can result in a degraded classification accuracy for SVM classifier at the testing time. Following their approach, Mei et al.~\cite{Mei2015using} introduced a more general framework for poisoning. Their method finds out an optimal change to the training set whenever the targeted learning model is trained using a convex optimization loss (e.g., SVMs or linear and logistic regression) and its input domain is continuous.
	\end{itemize}
	
	\subsubsection{Testing Phase Generation}
	\begin{itemize}
		
		\item \textit{White-Box Attacks}\label{sec:white-box}
		In this subsection, we precisely discuss how adversaries craft adversarial samples in a white-box setup. Papernot et al.~\cite{papernot2016distillation} introduced a general framework which builds on the attack approaches discussed in recent literature. The framework is split into two phases: a) \emph{direction sensitivity estimation} and b) \emph{perturbation selection} as shown in Figure~\ref{fig:adversarial_example}. The figure proposes an adversarial example crafting process for an image classification using DNN, which can be generalized for any supervised learning algorithm.
		
		\begin{figure}[!t]
			\centering
			\includegraphics[width=\textwidth]{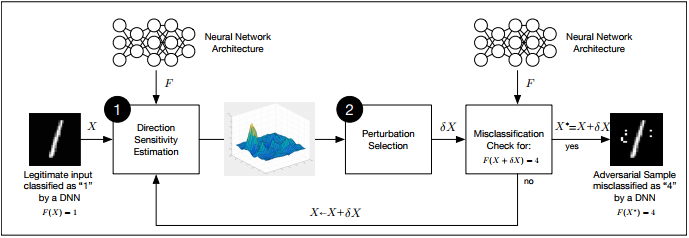}
			\caption{Adversarial Example Crafting Framework for Evasion Attacks~\label{fig:adversarial_example}}
			\vspace{-3mm}
			\caption*{(Image Credit: Papernot et al. \cite{papernot2016distillation})}
		\end{figure}
		
		Suppose $X$ is an input sample, and $F$ is a trained DNN classification model. The objective of an adversary is to craft a malicious example $X_{*} = X + \delta X$ by adding a perturbation $\delta X$ with the sample $X$, so that $F(X_{*}) = Y_{*}$ where $Y_{*} \neq F(X)$ is the target output which depends on the objective of the adversary. An adversary begins with a legitimate sample $X$. Since the attack setting is a white-box attack, the capability of an adversary is limited to accessing parameters $\theta$ of the targeted model $F$. The adversary employs a two-step process for the adversarial sample crafting, which is discussed below:
		
		\begin{enumerate}
			\item \textit{\textbf{Direction Sensitivity Estimation:}} The adversary evaluates the sensitivity of a class change to each input feature by identifying directions in the data manifold around sample $X$ in which the model $F$, learned by the DNN is most sensitive and likely to result in a class change.
			\item \textit{\textbf{Perturbation Selection:}} The adversary then exploits the knowledge of sensitive information to select a perturbation $\delta X$ among the input dimensions in order to obtain an adversarial perturbation which is most efficient.
		\end{enumerate}
		
		Both the steps are repeated by replacing $X$ with $X+\delta X$ before the start of each new iteration, until the adversarial goal is satisfied by the perturbed sample. The point to be remembered in this context is that the total perturbation used for crafting the adversarial sample from a valid example needs to be as minimum as possible. This is necessary for the adversarial samples to remain undetected in human eyes.
		
		Adversarial sample crafting using large perturbations is trivial. Thus, if one defines a norm $\|.\|$ to appropriately describe the differences between points in the input domain of the DNN model $F$, we can formalize the adversarial samples as a solution to the following optimization problem:
		
		\begin{equation}
		\label{eq:adv_example}
		X_{*} = X + \argmin\limits_{\delta X} \{\|\delta X\|: F(X+\delta X) \neq F(X) \}
		\end{equation}
		
		Most DNN models make this formulation non-linear and non-convex, making it hard to find a closed-solution in most of the cases. We now describe in details different techniques to find an approximate solution to this optimization problem using each of the two steps mentioned above.
		
		\subsubsection*{Direction Sensitivity Estimation}
		In this step, the adversary considers a legitimate sample $X$, an $n$-dimensional input vector. The objective here is to find those dimensions of $X$ which will produce an expected adversarial performance with the smallest selected perturbation. This can be achieved by changing the input components of $X$ and evaluating the sensitivity of the trained DNN model $F$ for these changes. Developing the knowledge of the model sensitivity can be accomplished in several ways. Some of the well-known techniques mentioned in the recent literature are discussed below.
		
		\begin{enumerate}
			\item \textbf{L-BFGS:} Szegedy et al.~\cite{szegedy2013intriguing} first introduced the term adversarial sample by formalizing the following minimization problem as the search for adversarial examples.
			
			\begin{equation*}
			\centering
			\argmin\limits_{r} f(x+r) = l \enspace \enspace \text{s.t.} \enspace (x+r) \in D
			\end{equation*}
			
			The input example $x$, which is correctly classified by f, is perturbed with $r$ to obtain the  resulting adversarial example $x_{*} = x + r$. The perturbed sample remains in the input domain $D$, however, it is assigned the target label $l \neq h(x)$. For non-convex models like DNN, the authors have used the \emph{L-BFGS}~\cite{liu1989limited} optimization algorithm to solve the above equation. Though the method gives good performance, it is computationally expensive while calculating adversarial samples.
			
			\item \textbf{Fast Gradient Sign Method (FGSM):} An efficient solution to equation~\ref{eq:adv_example} is introduced by Goodfellow et al.~\cite{goodfellow2014explaining}. They proposed a fast gradient sign methodology which calculates the gradient of the cost function with respect to the input of the neural network. The adversarial examples are generated using the following equation:
			
			\begin{equation*}
			X_{*} = X + \epsilon*sign(\nabla_x J(X, y_{true}))
			\end{equation*}
			
			Here, $J$ is the cost function of the trained model, $\nabla_x$ denotes the gradient of the model with respect to a normal sample $X$ with correct label $y_{true}$, and $\epsilon$ denotes the input variation parameter which controls the perturbation's amplitude. Recent literature have used some other variations of FGSM, which are summarised as below:
			
			\begin{enumerate}
				\item \textbf{Target Class Method:} This variant of FGSM~\cite{kurakin2016adversarial} approach maximizes the probability of some specific target class $y_{target}$, which is unlikely the true class for a given example. The adversarial example is crafted using the following equation:
				
				\begin{equation*}
				X_{*} = X - \epsilon*sign(\nabla_x J(X, y_{target}))
				\end{equation*}
				
				\item \textbf{Basic Iterative Method:} This is a straightforward extension of the basic FGSM method~\cite{kurakin2016adversarial}. This method generates adversarial samples iteratively using small step size.
				
				\begin{equation*}
				X_{*}^{0} = X ;\enspace \enspace X_{*}^{n+1} = Clip_{X, e} \{ X_{*}^{n} + \alpha*sign(\nabla_x J(X_{*}^{n}, y_{true}))\}
				\end{equation*}
				
				Here, $\alpha$ is the step size and $Clip_{X, e} \{A\}$ denotes the element-wise clipping of $X$. The range of $A_{i, j}$ after clipping belongs in the interval $[X_{i, j} - \epsilon, X_{i, j} + \epsilon]$. This method does not typically rely on any approximation of the model and produces additional harmful adversarial examples when run for more iterations.
			\end{enumerate}
			
			\item \textbf{Jacobian Based Method:} Papernot et al.~\cite{papernot2016limitations} introduced a different approach for finding sensitivity direction by using forward derivative, which is the Jacobian of the trained model $F$. This method directly provides gradients of the output components with respect to each input component. The knowledge thus obtained is used to craft adversarial samples using a complex saliency map approach which we will discuss later. This method is particularly useful for source-target misclassication attacks.
		\end{enumerate}
		
		\subsubsection*{Perturbation Selection}
		An adversary may use the information about network sensitivity for input differences in order to evaluate the dimensions which are most likely to generate the target misclassification with minimum perturbation. The perturbed input dimensions can be of two types:
		
		\begin{enumerate}
			\item \textbf{Perturb all the input dimensions:} Goodfellow et al.~\cite{goodfellow2014explaining} proposed a way to perturb every input dimensions but with a small quantity in the direction of the sign of the gradient calculated using the FGSM method. This method efficiently minimizes the Euclidian distance between the original and the corresponding adversarial samples.
			
			\item \textbf{Perturb a selected input dimensions:} Papernot et al.~\cite{papernot2016limitations} choose to follow a more complicated process involving saliency maps to select only a limited number of input dimensions to perturb. The objective of using saliency map is to assign values to the combination of input dimensions which indicates whether the combination if perturbed, will contribute to the adversarial goals. This method effectively reduces the number of input features perturbed while crafting adversarial examples. For choosing the input dimensions which forms the perturbations, all the dimensions are sorted in decreasing order of adversarial saliency value. The saliency value $S(x, t)[i]$ of a component $i$ of a legitimate example $x$ for a target class $t$ is evaluated using the following equation:
			
			\begin{equation*}
			S(x, t)[i] =
			\begin{cases}
			0, & \text{if}\ \frac{\partial F_t}{\partial x_{i}} (x) < 0 \enspace \text{or} \enspace \sum_{j \neq t}^{}\frac{\partial F_j}{\partial x_{i}} (x) > 0\\
			\frac{\partial F_t}{\partial x_{i}} (x)\left|\sum_{j \neq t}^{}\frac{\partial F_j}{\partial x_{i}} (x)\right|, & \text{otherwise}
			\end{cases}
			\end{equation*}
			
			where $\left[ \frac{\partial F_j}{\partial x_{i}} \right]_{ij}$ could be easily calculated using the Jacobian Matrix $J_{F}$ of the learned model $F$. Input components are added to perturbation $\delta x$in the decreasing order of the saliency value $S(x, t)[i]$ until the resulting sample $x_{*} = x + \delta x$ is misclassified by the model $F$.
			
		\end{enumerate}
		
		Each method has its own advantages and drawbacks. The first method is well fitted for the fast crafting of many adversarial samples but with a relatively large perturbation and thus is potentially easier to detect. The second method reduces the perturbations at the expense of a higher computing cost.
		
		\item \textit{Black-Box Attacks}
		In this subsection, we discuss in details on the generation of adversarial examples in a black-box setting. Crafting adversarial samples in non-adaptive and strict black box scenario is straightforward. In both the cases, the adversary has access to a vast dataset to train a local substitute model which approximates the decision boundary of the target model. Once the local model is trained with high confidence, any of the white-box attack strategies can be applied on the local model to generate adversarial examples, which eventually can be used to fool the target model because of the \emph{Transferability Property of Neural Network} (will be discussed later). However, in the case of an adaptive black-box scenario, the adversary does not have access to a large dataset and thus augments a partial or randomly selected dataset by selectively querying the target model as an oracle. One of the popular method of dataset augmentation presented by Papernot et al.~\cite{papernot2017practical} is discussed next.
		
		\subsubsection*{Jacobian based Data Augmentation}
		An adversary could potentially make an infinite number of queries to get the Oracle's output $O(x)$ for any input $x$. This would provide the adversary a copy of the oracle. However, the process is not tractable considering the continuous domain of an input to be queried. Furthermore, making a significant number of queries presents the adversarial behavior easy to detect. The heuristic that can be used to craft synthetic training inputs is based on identifying the directions in which the target model\textquotesingle s output is varying, around an initial set of training points. With more input-output pair the direction can be captured easily for a target Oracle $O$. Hence, the greedy heuristic that an adversary follow is to prioritize the samples while querying the oracle for labels to get a substitute DNN $F$ approximating the decision boundaries of the Oracle. These directions can be identified with the substitute DNN\textquotesingle s Jacobian matrix $J_F$, which is evaluated at several input points $x$. Precisely, the adversary evaluates the sign of the Jacobian matrix dimension corresponding to the label assigned to input $x$ by the oracle, denoted by $sgn\left(J_F(x)\left[O(x)\right] \right)$. The term $\lambda * sgn\left(J_F(x)\left[O(x)\right] \right)$ is added to the original datapoint, to obtain a new synthetic training point. The iterative data augmentation technique can be summarized using the following equation.
		
		\begin{equation*}
		S_{n+1} = \{x + \lambda * sgn\left(J_F(x)\left[O(x)\right] \right) : x \in S_n\} \cup S_n
		\end{equation*}
		
		where $S_n$ is the dataset at $n^{th}$ step and $S_{n+1}$ is the augmented dataset. The substitute model training following this approach is presented in Figure~\ref{fig:jac_aug}.
		
		\begin{figure}[!t]
			\centering
			\includegraphics[width=\textwidth]{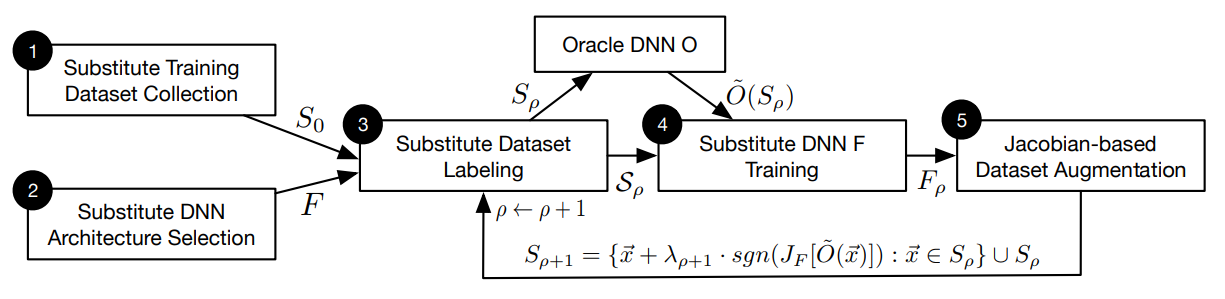}
			\caption{Substitute Mode Training using Jacobian based dataset augmentation.}\label{fig:jac_aug}
			\vspace{-3mm}
			\caption*{(Image Credit: Papernot et al. \cite{papernot2017practical})}
		\end{figure}
		
		\subsubsection{Transferability of Adversarial Samples}
		Adversarial sample transferability \cite{papernot2016transferability} is the property that adversarial samples produced by training on a specific model can affect another model, even if they have different architectures - leading to adaptive black box attacks as discussed in Section \ref{black-box}. Since in case of black-box attack, adversary does not have access to the target model $F$, an attacker can train a \emph{substitute model} $F'$ locally to generate adversarial example $X + \delta X$ which then can be transfered to the victim model. Formally, if $X$ is the original input, the transferability problem can be represented as an optimization problem \eqref{eq:adv_example}.
		
		It can be broadly classified into two types:
		\begin{enumerate}
			\item Intra-technique transferability:
			If models $F$ and $F'$ are both trained using same machine learning technique (e.g. both are NN or SVM)
			\item Cross-technique transferability: If learning technique in $F$ and $F'$ are different, for example, $F$ is a neural network and $F'$ is a SVM.
		\end{enumerate}
		
		This substitute model in effect transfers knowledge crafting adversarial inputs to the victim model. The targeted classifier is designated as an oracle because
		adversaries have the minimal capability of querying it for predictions on inputs of their choice. To train the substitute model, dataset augmentation as discussed above is used to capture more information about the predicted outputs. Authors also introduces reservoir sampling to select a limited number of new inputs while performing Jacobian-based data augmentation, reduces the number of queries made to the oracle. The choice of substitute model architecture has limited impact on transferability.
		
		The attacks have been shown to generalize to non-differentiable target models like decision trees, e.g, deep neural networks (DNNs) and logistic regression (LR) (differentiable models) could both effectively be used to learn a substitute model for many classifiers trained with a support vector machine, decision tree, and nearest neighbor. Cross-technique transferability reduces the amount of knowledge that adversaries must possess in order to force misclassification by crafted samples. Learning substitute model alleviates the need of attacks to infer architecture, learning model and parameters in a typical black-box based attack.
	\end{itemize}
	
    \subsection{GAN based attack in Collaborative Deep Learning}
	    { Hitaj et al. \cite{hitaj2017deep} presented a GAN based attack to extract information from honest victims in a collaborative deep learning framework. The goal of GAN is to produce samples identical to those in the training set without having access to the original training set. GAN-based method works only during the training phase in collaborative deep learning. The authors showed that the attack can be carried out in Convolutional Neural Networks which are themselves pretty difficult to invert or even when the parameters are hidden using differential privacy. In the white-box setting, the adversary acts as an insider within the privacy-preserving \cite{shaham2015understanding} collaborative deep learning protocol. The motive of the adversary is to extract tangible information about the labels that do not belong to his dataset.
	
	\begin{figure}[!t]
		\centering
		\includegraphics[width=\textwidth]{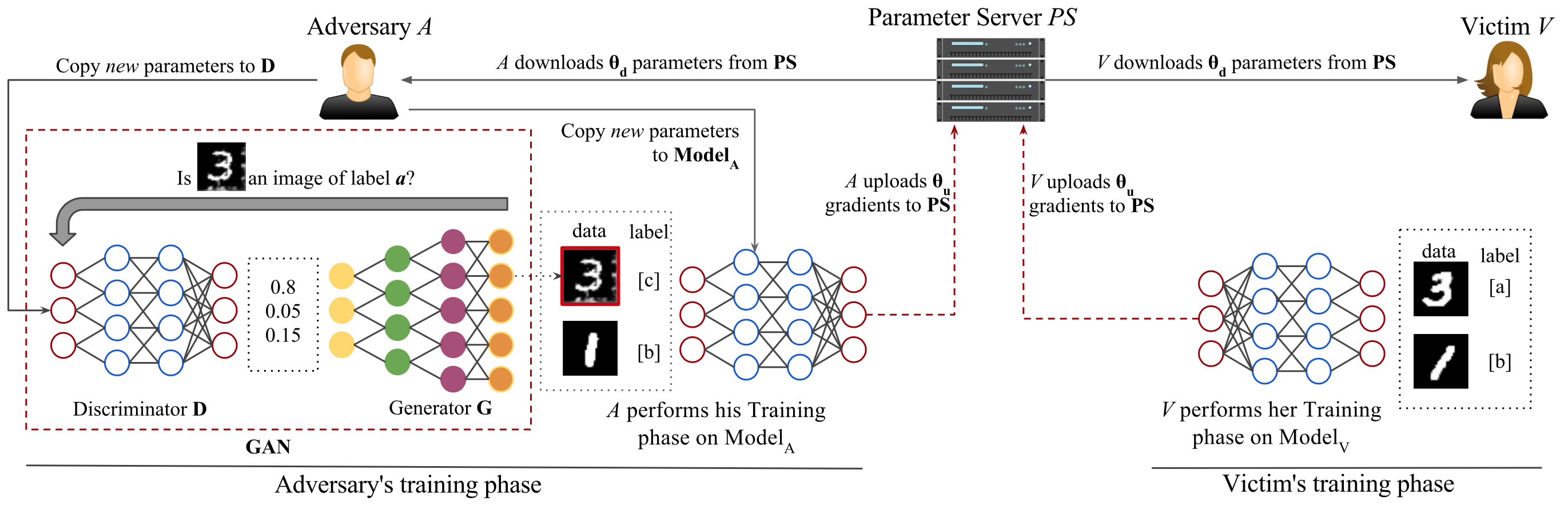}
		\caption{GAN Attack on collaborative deep learning}
		\vspace*{-3mm}
		\caption*{(Image Credit: Hitaj et al. \cite{hitaj2017deep})}
		\label{fig:gan_coll}
	\end{figure}
	
	Figure \ref{fig:gan_coll} shows the proposed attack, which uses GAN to generate similar samples as that of training data in a collaborative learning setting, with limited access to shared parameters of the model. 
	$A$, $V$ participates in a collaborative learning. V(the victim) chooses labels $[x, y]$. The adversary $A$ also chooses labels $[y, z]$. Thus, while class $y$ is common to both $A$ and $V$,  $A$ does not have any information about the $x$. The goal of the adversary is to collect maximum information possible about the class $x$. Using GAN, the adversary generates samples which are identical to class $x$. The insider injects these fake samples from $x$, as class $z$ into the distributed learning procedure. The victim unable to distinguish between classes $x$ and $z$ reveals more information about class $x$ than initially intended. Thus, the insider imitates samples from $x$ and uses the victim to improve his knowledge about a class he ignored before training. The GAN attack works as long as $A$\textquotesingle s local model improves its accuracy over time.
	
	Authors \cite{hitaj2017deep} proved that GAN attack was successful in all set of experiments conducted juxtaposed with MI	attacks, DP based collaborative learning, etc. The GAN will generate good samples as long as the discriminator is learning. }
	
	\subsection{Adversarial Classification}
	{ Dalvi et al \cite{dalvi2004adversarial} defined adversarial classification from cost-sensitive game-theoretical perspective as a game between Classifier, who tries to learn from training set a function $y_c$ = $C$($x$) that will accurately predict the classes of instances in training set, and Adversary, who tries to make Classifier predict positive instance of the training set as negative by modifying those instances from $x$ to $x$\textquotesingle = $A$($x$). The classifier and the adversary are constantly trying to defeat each other by maximizing their own payoffs. The classifier uses a cost-sensitive Bayes learner to minimize its expected cost while assuming that the adversary always plays its optimal strategy, whereas the adversary tries to modify feature in order to minimize its own expected cost \cite{barreno2010security} \cite{zhou2018gametheory}. The presence of adaptive adversaries in a system can significantly degrade the performance of the classifier, specially when the classifier is unaware of the presence or type of adversary. 
	The goal of the Classifier is to build a classifier $C$ to maximize its expected utility ($U_C$), whereas the goal of the Adversary is to find a feature change strategy $A$ in order to maximize its own expected utility:
	\begin{equation*}
	    U_C = \sum_{(x,y) \epsilon XY} P(x,y) \: \bigg[ U_{C}(C(A(x),y) - \sum_{X_{i} \epsilon X_{C}(x)} V_{i} \bigg]
	\end{equation*}
	
	\begin{equation*}
	    U_A = \sum_{(x,y) \epsilon XY} P(x,y) \: [ U_{A}(C(A(x),y) -  W(x,A(x)) ]
	\end{equation*}
	
	}
	
	\subsection{Evasion and Poisoning attack on Support Vector Machines}
	{ Support Vector Machines (SVM) are one of the most popular classification techniques widely used for malware detection,. intrusion detection systems and spam filtering. SVM follows \textit{stationarity} property, i.e., both the training and test data should come from the same distribution. However, in adversarial learning, an intelligent and adaptive adversary can manipulate data thereby violating stationarity to exploit the vulnerabilities of the learning system. 
	
	Biggio et al \cite{biggio2014secevaluation} using gradient-descent based approach demonstrated the evasion attack on kernel-based classifiers \cite{golland2001discriminative}. Even if the adversary is unaware of the classifier's decision function, he can learn a \textit{surrogate} classifier to evade the classifier. The authors considered two approaches: in the first, the adversary had knowledge about feature space and classifier's discriminant function, while in the second, the adversary did not know the learned classifier. In the second scenario, the adversary is assumed to learn a \textit{surrogate} classifier on a surrogate training set. So, even if the adversary can learn a classifier's copy on surrogate data, SVMs can be evaded with relatively high probability. So, the optimal strategy to find an attack sample $x$ that will minimize the value of the classifier's discriminant function $g(x)$ can be given as:
	\begin{equation*}
    			x^{*} = \argmin_{x} \hat{g}(x) \qquad s.t \quad d(x,x^0) \leq d_{max}
	\end{equation*}
    However, if \^{g}(x) is not convex, the gradient descent may lead to a local minima outside of sample's support. To overcome this problem, additional components were introduced into the equation yielding the following modified equation:
    \begin{equation*}
        \argmin_{x} E(x) = \hat{g}(x) - \frac{\lambda}{n} \sum_{i|f_{i} = -1} k(\frac{x - x_i}{h}) \qquad s.t \quad d(x,x^0) \leq d_{max}
    \end{equation*}

	Biggio et al \cite{Biggio2012poisoning} also demonstrated poisoning attacks against SVMs in which the adversary manipulates training data to force the SVM to misclassify test samples. Although the training data was tampered to include well-crafted attack samples, it was assumed that there was no manipulation of test data. For the attack, the adversary's goal is to find a set of points whose addition to the training dataset will maximally decrease the SVM's classification accuracy. Although having complete knowledge of the training dataset is difficult in real-world scenarios, but collecting a \textit{surrogate} dataset having the same distribution as the actual training dataset may not be difficult for the attacker. Under these assumptions, the optimal attack strategy can be given as:
	\begin{equation*}
	    x^* = \argmax_{x} P(x) = \sum_{k=1}^{m} (1 - y_{k}f_{x}(x_k))_{+} = \sum_{k=1}^{m} (-g_{k})_{+} \qquad s.t \quad x_{lb} \leq x \leq x_{ub}
	\end{equation*}

	Another class of attacks on SVM called privacy attack, presented by Rubinstein et al \cite{rubinstein2009privacysvm}, is based on the attempt to breach of training data's confidentiality. The ultimate goal of the adversary is to determine features and/or the label of an individual training instance by trying to inspect test-time classifications made by the classifier or inspecting the classifier directly.
	}
	
	\subsection{Poisoning attacks on Collaborative Systems}
	{
	Recommendation and collaborating filtering systems play an important part in the business strategies of modern e-commerce systems. The performance of these systems can seriously affect the business, both in a positive and negative way, thereby making them attractive targets for the adversaries. Bo Li et al \cite{Li2016data} demonstrated poisoning attacks on collaborative filtering systems where an attacker, having complete knowledge of the learner, can generate malicious data to degrade the effectiveness of the system. 
	
	The two most popular algorithms used for factorization-based collaborative filtering are alternating minimization \cite{jain2012altmin} and nuclear norm minimization \cite{cai2010matcomp}. For the former one, the data matrix $M$ can be determined by the following optimization problem:

	\begin{equation*}
	    \min_{{U \epsilon \mathbb{R}^{m \times k}} , {V \epsilon \mathbb{R}^{n \times k}}} \big\{ {\left \| \mathcal{R}_{\Omega} (M - UV^{T}) \right \|}^2_F + 2 \lambda_{U} {\left \| U \right \|}^2_F + 2 \lambda_{V} {\left \| V \right \|}^2_F \big \} \tag{1}  \label{eq:1}
	\end{equation*}
	
	Alternatively, the latter one can be solved by the following problem:
	\begin{equation*}
	    \min_{{X \epsilon \mathbb{R}^{m \times k}}} \big\{ {\left \| \mathcal{R}_{\Omega} (M - X) \right \|}^2_F + 2 \lambda {\left \| X \right \|}_*  \big \} \tag{2}  \label{eq:2}
	\end{equation*}

	where $\lambda$ > 0 is a regularization parameter and  ${\left \| X \right \|}_* \ = \sum_{i=1}^{rank(X)} | \sigma_i (X) |$  is the nuclear form of $X$ .

	Based on these problems, the authors Bo Li et al \cite{Li2016data} listed the three types of attacks and their utility functions:
	\begin{itemize}
	\item \textit{Availability Attack :} In this attack model, the goal of the attacker is to maximize the error of the collaborative filtering system thereby making it unreliable and useless. The utility function can be defined as the total number of perturbations of predictions between $\overline{M}$ (prediction without data poisoning) and $\widehat{M}$ (prediction after poisoning) on unseen entries $\Omega^C$
	\begin{equation*}
	    R^{av} (\widehat{M}, M) = {\left \| \mathcal{R}_{\Omega_C} (\widehat{M} - \overline{M} ) \right \| }^2_F
	\end{equation*}
	
	\item \textit{Integrity Attack :} In this model, the attacker tries to manipulate the popularity of a subset of items. Thus, if $J_0$ $\subseteq$ [n] is the subset of items and $w: J_{0} \rightarrow \mathbb{R} $ is a pre-specified weight vector, then the utility function can be defined by
	\begin{equation*}
	    R^{in}_{J_0 , w} (\widehat{M}, M) = \sum_{i = 1}^m \sum_{j \epsilon J_0} w(j) \widehat{M}_{ij}
	\end{equation*}
	    
	 \item \textit{Hybrid Attack :} It is a hybrid of the above two models, defined by the function
	 \begin{equation*}
	     R^{hybrid}_{J_0, w, \mu} (\widehat{M}, M) = \mu_1 R^{av}(\widehat{M}, M) + \mu_2 R^{in}_{J_0 , w} (\widehat{M}, M) 
	 \end{equation*}
	 where $\mu = (\mu_1, \mu_2)$ are the coefficients that provides the trade off between availability and integrity attacks.
    \end{itemize}
	}
	
	\subsection{Adversarial attacks on Anomaly Detection Systems}
	{
	Anomaly detection is referred to the detection of events that do not conform to an expected pattern or behaviour. Kloft et al \cite{kloft2010online} analyzed the behaviour of online centroid anomaly detection systems under poisoning attack. The purpose of anomaly detection system is to determine whether a test example $x$ is likely to originate from the same distribution as the given dataset $X$. It flags $x$ as outlier if it lies in a region of low density compared with the probability density function of sample space $X$. The authors used \textit{finite sliding window} of training data where, with the arrival of every new data point, the centre of mass changes by 
	\begin{equation*}
	    	c\textquotesingle = c + {{1 \over{n}}{(x - x_i)}}
	\end{equation*}

    As illustrated in Figure.~\ref{fig:online_poison}, the adversary will try to force the anomaly detection algorithm to accept an attack point $A$ which lies outside the normal ball, i.e. $||A - c|| > r$. 
    
    \begin{figure}[!h]
		\centering
		\includegraphics [scale=0.66] {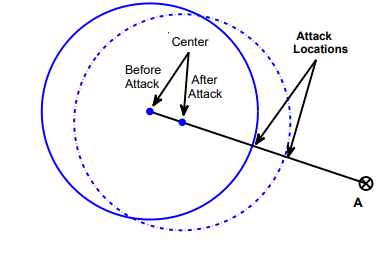}
		\caption{Illustration of poisoning attack}
		\vspace*{-3mm}
		\caption*{(Image Credit: Kloft et al. \cite{kloft2010online})}
		\label{fig:online_poison}
	\end{figure}
	}
	
	Although the attacker is assumed to have knowledge about the algorithm and the training data, he cannot modify existing data other than adding new points.

	\section{Advances in Defense Strategies} \label{sec:advancements}
	In this section, we provide a brief discussion about the recent advancements in defense strategies. Adversarial examples demonstrate that many modern machine learning algorithms can be broken easily in surprising ways. A substantial amount of research to provide a practical defense against these adversarial examples can be found in recent literature. These adversarial examples are hard to defend because of the following reasons~\cite{online:openai}:
	
	\begin{enumerate}
		\item \emph{A theoretical model of the adversarial example crafting process is very difficult to construct.} The adversarial sample crafting is a complex optimization process which is non-linear and non-convex for most Machine Learning models. The lacking of proper theoretical tools to describe the solution to these complex optimization problems make it even harder to make any theoretical argument that a particular defense will rule out a set of adversarial examples.
		\item \emph{Machine Learning models are required to provide proper outputs for every possible input.} A considerable modification of the model to incorporate robustness against the adversarial examples may change the elementary objective of the model.
	\end{enumerate}
	
	Most of the current defense strategies are not adaptive to all types of adversarial attack as one method may block one kind of attack but leaves another vulnerability open to an attacker who knows the underlying defense mechanism. Moreover, implementation of such defense strategies may incur performance overhead, and can also degrade the prediction accuracy of the actual model. In this section, we discuss the most recent defense mechanisms used to block these adversarial attacks along with their respective possible drawbacks of implementation.
	
	The existing defense mechanisms can be categorized among the following types based on their methods of implementation.
	
	\subsection{Adversarial Training}
	
	The primary objective of the adversarial training is to increase model robustness by injecting adversarial examples into the training set~\cite{szegedy2013intriguing, goodfellow2014explaining, shaham2015understanding, lyu2015aunified}. Adversarial training is a standard brute force approach where the defender simply generates a lot of adversarial examples and augments these perturbed data while training the targeted model. The augmentation can be done either by feeding the model with both the original data and the crafted data, presented in ~\cite{kurakin2016adversarial} or by learning with a modified objective function given by~\cite{goodfellow2014explaining} -
	
	\begin{equation*}
	\widetilde{J}(\theta, x, y) = \alpha J(\theta, x, y) + (1-\alpha)J(\theta, x + \epsilon sign(\nabla_x J(\theta, x, y)), y)
	\end{equation*}
	
	with $J$ being the original loss function. The central idea behind this strategy is to increase model\textquotesingle s robustness by ensuring that it will predict the same class for legitimate as well as perturbed examples in the same direction.
	Traditionally, the additional instances are crafted using one or multiple attack strategies mentioned in Section~\ref{sec:white-box}.
	
	Adversarial training of a model is useful only on adversarial examples which are crafted on the original model. The defense is not robust for black-box attacks~\cite{papernot2017practical, narodytska2017simple} where an adversary generates malicious examples on a locally trained substitute model. Moreover, Tram\`er et al.~\cite{Tramer2017ensemble} have already proved that the adversarial training can be easily bypassed through a two-step attack, where random perturbations are applied to an instance first and then any classical attack technique is performed on it, as mentioned in Section~\ref{sec:white-box}.
	
	\subsection{Gradient Hiding}
	A natural defense against gradient-based attacks presented in~\cite{Tramer2017ensemble} and attacks using adversarial crafting method such as FGSM, could consist in hiding information about the model's gradient from the adversary. For instance, if the model is non-differentiable (e.g, a Decision Tree, a Nearest Neighbor Classifier, or a Random Forest), gradient-based attacks are rendered ineffective. However, this defense are easily fooled by learning a surrogate Black-Box model having gradient and crafting examples using it~\cite{papernot2017practical}.
	
	\subsection{Defensive Distillation}
	The term distillation was originally proposed by Hinton et al.~\cite{hinton2015distillating} as a way to transfer knowledge from a large neural networks to a smaller one. Papernot et al.~\cite{papernot2016distillation, papernot2017extending} recently proposed using it as a defensive mechanism against adversarial example crafting methods.
	
	The two-step working of the distillation method is described as below. Let us assume that we already have a neural network $F$ which classifies a training dataset $X$ into the target classes $Y$. The final softmax layer in the produces a probability distribution over $Y$. Further, assume that we want to train a second neural network $F^{\prime}$ on the same dataset $X$ achieving the same performance. Now, instead of using the target class labels of the training dataset, we use the output of the network $F$ as the label to train another neural network $F^{\prime}$ with the same architecture having the same input dataset $X$. The new labels, therefore, contain more information about the membership of $X$ to the different target classes, compared to a simple label that just chooses the most likely class.
	
	The final softmax layer in the distillation method is modified according to the following equation:
	
	\begin{equation*}
	F_{i}(X) = \frac{e^{\frac{z_i (X)}{T}}}{\sum_{i=1}^{\left|Y\right|}e^{\frac{z_i (X)}{T}}}
	\end{equation*}
	
	where $T$ is the distillation parameter called temperature. Papernot et al.~\cite{papernot2016distillation} showed experimentally, a high empirical value of $T$ gives a better distillation performance. The advantage of training the second model using this approach is to provide a smoother loss function, which is more generalized for an unknown dataset and have high classification accuracy even for adversarial examples.
	
	Papernot et al. \cite{papernot2016distillation} used defensive distillation, learned by the DNN architecture, to smooth the model. The technique is known as label smoothing which involves converting class labels into soft targets. A value close to 1 is assigned for the target class and the rest of the weight is distributed to the other classes. These new values are used as labels for training the model instead of the true labels. As an outcome, the need to train an additional model as for defensive distillation is relaxed. The advantage of using soft targets lies in the fact that it uses probability vectors instead of hard class labels. For example, given an image $X$ of handwritten digits, the probability of similar looking digits will be close to one another. 
	
	However with the recent advancement in the black-box attack, the defensive distillation method as well as the label smoothing method can easily be avoided~\cite{papernot2017practical, carlini2017towards}. The main reason behind the success of these attacks is often the strong transferability of adversarial examples across neural network models.
	
	\subsection{Feature Squeezing}
	Feature squeezing is another model hardening technique~\cite{xu2017feature, xu2017feature2}. The main idea behind this defense is that it reduces the complexity of representing the data so that the adversarial perturbations disappear because of low sensitivity. There are mainly two heuristics behind the approach considering an image dataset.
	
	\begin{enumerate}
		\item Reduce the color depth on a pixel level, i.e., encoding the colors with fewer values.
		\item Use of a smoothing filter over the images. As a result, multiple inputs are mapped into a single value which makes the model safe against noise and adversarial attacks.
	\end{enumerate}
	
	Though these techniques work well in preventing adversarial attacks, these have the collateral effect of worsening the accuracy of the model on true examples
	
	\subsection{Blocking the Transferability}
	The main reason behind defeat of most of the well-known defense mechanism is due to the strong transferability property in the neural networks, i.e., adversarial examples generated on one classifier are expected to cause another classifier to perform the same mistake. The transferability property holds true even if the classifiers have different architectures or trained on disjoint datasets. Hence, the key for protecting against a black-box attack is to block the transferability of the adversarial examples.
	
\begin{figure}[!t]
		\centering
		\includegraphics[width=0.75\linewidth]{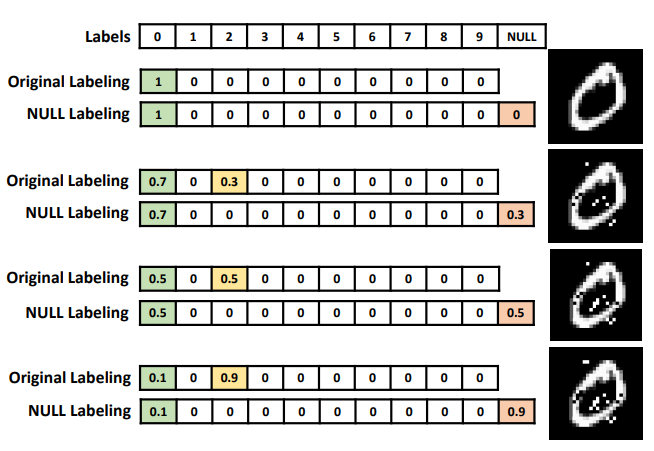}
		\caption{Illustration of NULL Labeling Method}
		\vspace{-3mm}
		\caption*{(Image Credit: Hosseini et al. \cite{hossain2017blocking})}
		\label{fig:null_label}
	\end{figure}

	Hosseini et al.~\cite{hossain2017blocking} recently proposed a three-step \emph{NULL Labeling} method to prevent the adversarial examples to transfer from one network to another. The main idea behind the proposed approach is to augment a new NULL label in the dataset and train the classifier to reject the adversarial examples by classifying them as NULL. The basic working of the approach is shown in Figure~\ref{fig:null_label}. The figure illustrates the method taking as an example image from MNIST dataset, and three adversarial examples with different perturbations. The classifier assigns a probability vector to each image. The NULL labeling method assigns a higher probability to the NULL label with higher perturbation, while the original labeling without the defense increases the probabilities of other labels.
	
	The NULL Labeling method is composed of three major steps:
	
	\begin{enumerate}
		\item \textbf{Initial Training of the target classifier:} Initial training is performed on the clean dataset to derive the decision boundaries for the classification task.
		\item \textbf{Computing the NULL probabilities:} The probability of belonging to the NULL class is then calculated using a function $f$ for the adversarial examples generated with different amount of perturbations.
		
		\begin{equation*}
		p_{NULL} = f\left(\frac{\|\delta X\|_0}{\|X\|}\right)
		\end{equation*}
		
		where, $\delta X$ is the perturbation and $\|\delta X\|_0 \sim U[1, N_{max}]$. $N_{max}$ is the minimum number for which $f(\frac{N_{max}}{\|X\|}) = 1$.
		
		\item \textbf{Adversarial Training:} Each clean sample is then re-trained with the original classifier along with different perturbed inputs for the sample. The label for the training data is decided based on the NULL probabilities obtained in the previous step.
	\end{enumerate}
	
	The advantage of this method is the labeling of the perturbed inputs to NULL label instead of classifying them into their original label. To the best of our knowledge, this method is the most effective defense against the adversarial attacks to date. This method is accurate to reject an adversarial example while not compromising the accuracy of the clean data.
	
	\subsection{Defense-GAN}
	{
	Samangouei et al. \cite{samangouei2018defensegan} proposed a mechanism to leverage the power of Generative Adversarial Networks \cite{goodfellow2014gan} to reduce the efficiency of adversarial perturbations, which works both for white box and black box attacks. In a typical GAN, a generative model which emulated the data distribution, and a discriminative model that differentiates between original input and perturbed input, are trained simultaneously. The central idea is to "project" input images onto the range of the generator $G$ by minimizing the reconstruction error ${\left \| {G(z) - x} \right \|}_{2}^{2} $ , prior to feeding the image $x$ to the classifier. Due to this, the legitimate samples will be close to the range of $G$ than the adversarial samples, resulting in substantial reduction of potential adversarial perturbations. An overview of the Defense-GAN mechanism is shown in Figure.~\ref{fig: defense_gan}.
	
	Although Defense-GAN showed to be quite effective against adversarial attacks, its success relies on the expressiveness and generative power of the GAN. Moreover, training of GAN can be challenging and if not properly trained, the performance of the Defense-GAN can significantly degrade.
	
	\begin{figure}[!h]
		\centering
		\includegraphics[width=\linewidth]{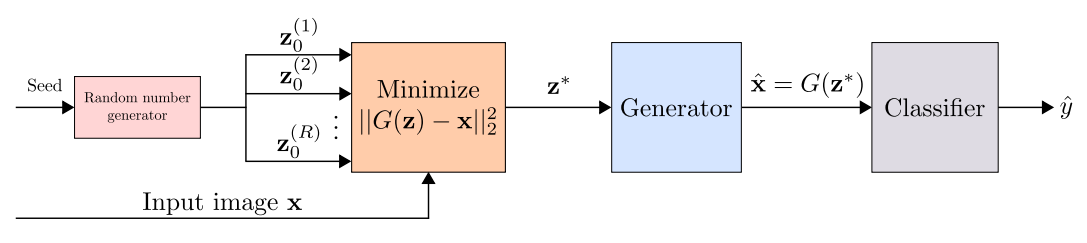}
		\caption{Overview of Defense-GAN algorithm}
		\vspace*{-3mm}
		\caption*{(Image Credit: Samangouei et al. \cite{samangouei2018defensegan})}
		\label{fig: defense_gan}
	\end{figure}
	}
	
	\subsection{MagNet}
	{
	Meng et al. \cite{meng2017magnet} proposed a framework, named MagNet, which uses classifier as a black box to read the output of the classifier's last layer without reading the data on any internal layer or modifying the classifier. MagNet uses \textit{detectors} to distinguish between a normal and adversarial example. The detector measures the distance between the given test example and the manifold and rejects it if the distance exceeds a threshold. It also uses a \textit{reformer} to reform adversarial example to a similar legitimate example using autoencoders. Although MagNet was successful in thwarting a range of black-box attacks, its performance degraded significantly in case of white-box attacks where the attackers are supposed to be aware of the parameters of MagNet. So, the authors came up with the idea of using varieties of autoencoders and randomly pick one at a time to make it difficult for the adversary to predict which autoencoder was used.
	
	\begin{figure}[!t]
		\centering
		\includegraphics [scale=0.45] {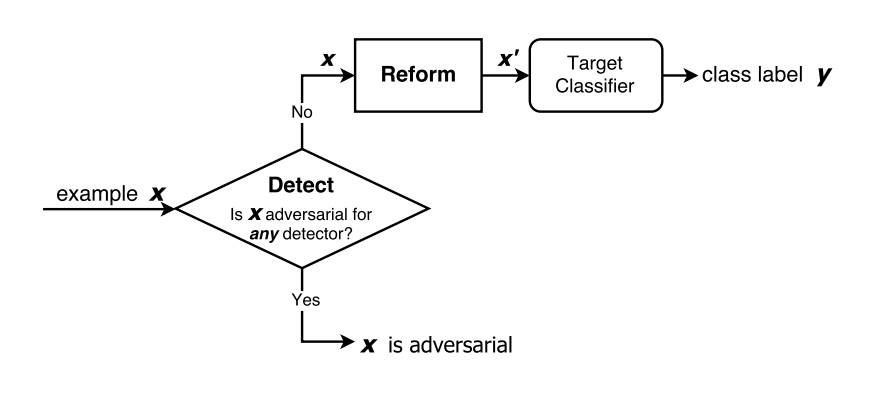}
		\caption{MagNet workflow in test phase.}
		\vspace*{-3mm}
		\caption*{(Image Credit: Meng et al. \cite{meng2017magnet})}
		\label{fig: magnet}
	\end{figure}
	
	}
	
	\subsection{Using High-Level Representation Guided Denoiser}
	{ While standard denoisers like pixel-level reconstruction loss function, suffer from error amplification, high-level representation guided denoiser(HGD) can effectively overcome this problem by using a loss function which compares the target model's output produced by clean image and denoised image. Liao et al \cite{liao2017hgd} introduced HGD to devise a robust target model immune against white-box and black-box adversarial attacks. Another advantage of using HGD is that it can be trained on a relatively small dataset and can be used to protect models other than the one guiding it. 
	
	In the HGD model, the loss function is defined as the $ L_1 $ norm of the difference between the $l$-th layer representation of the neural network, activated by $x$ and $\hat{x}$
	\begin{equation*}
	    L = \left \| f_l(\hat{x} - f_l(x) \right \|
	\end{equation*}
	The authors \cite{liao2017hgd} proposed three different training methods for HGD, illustrated in Figure.~\ref{fig: hgd}. In FGD (Feature Guided Denoiser), they defined $l = -2$ as the index for the topmost convolutional layer and fed the activations of that layer to the linear classification layer after global average pooling. In LGD (Logits Guided Denoiser), they defined $l = -1$ as the index for the logits, i.e., the layer before the softmax layer. While both of these variants were unsupervised models, they proposed a supervised variant, CGD (Class label Guided Denoiser), which uses classification loss of the target model as the loss function.
	
	\begin{figure}[!t]
		\centering
		\includegraphics [width=\linewidth] {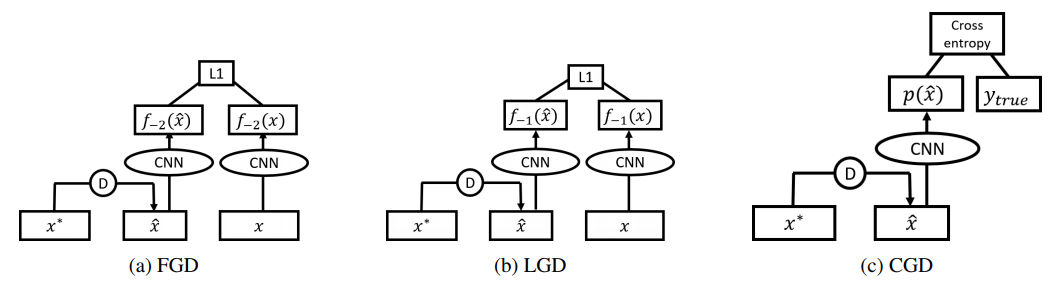}
		\caption{Three different training methods for HGD. D stands for denoiser, CNN is the model to the defended.}
		\vspace*{-3mm}
		\caption*{(Image Credit: Liao et al. \cite{liao2017hgd})}
		\label{fig: hgd}
	\end{figure}
	}
	
	\subsection{Using Basis Function Transformations}
	{
	Shaham et al. \cite{shaham2018basisfunc} investigated various defense mechanisms like PCA, low-pass filtering, JPEG compression, soft thresholding, etc. by manipulations based on basis function representation of images. All the mechanisms were applied as a pre-processing step on both adversarial and legitimate images and evaluated the efficiency of each technique by their success at distinguishing between the two sets of images. The authors showed that JPEG compression performs better than all the other defence mechanisms under consideration across all types of adversarial attacks in black-box, grey-box and white-box settings.
	}

	As described, the existing defense mechanisms have their limitations in the sense that they can provide robustness against specific attacks in specific settings. The design of a robust machine learning model against all types of adversarial examples is still an open research problem.
	
	\section{Conclusion}\label{sec:conclusion}
	{
	Despite their high accuracy and performance, machine learning algorithms have been found to be vulnerable to subtle perturbations that can have catastrophic consequences in security related environments. The threat becomes more grave when the applications operate in adversarial environment. So, it has become immediate necessity to devise robust learning techniques resilient to adversarial attacks.	A number of research papers on adversarial attacks as well as their countermeasures has surfaced since Szegedy et al \cite{szegedy2013intriguing} demonstrated the vulnerability of machine learning algorithms. In this paper we have tried to explore some of the well known attacks and proposed defense strategies. We have also tried to provide a taxonomy on topics related to adversarial learning. After the review we can conclude that adversarial learning is a real threat to application of machine learning in physical world. Although there exists certain countermeasures, but none of them can act as a panacea for all challenges. It remains as an open problem for the machine learning community to come up with a considerably robust design against these adversarial attacks.
	}
	
\bibliographystyle{ACM-Reference-Format}
\bibliography{ref_ml}

%% file: sample-acmlarge.bbl

\begin{thebibliography}{79}


\ifx \showCODEN    \undefined \def \showCODEN     #1{\unskip}     \fi
\ifx \showDOI      \undefined \def \showDOI       #1{#1}\fi
\ifx \showISBNx    \undefined \def \showISBNx     #1{\unskip}     \fi
\ifx \showISBNxiii \undefined \def \showISBNxiii  #1{\unskip}     \fi
\ifx \showISSN     \undefined \def \showISSN      #1{\unskip}     \fi
\ifx \showLCCN     \undefined \def \showLCCN      #1{\unskip}     \fi
\ifx \shownote     \undefined \def \shownote      #1{#1}          \fi
\ifx \showarticletitle \undefined \def \showarticletitle #1{#1}   \fi
\ifx \showURL      \undefined \def \showURL       {\relax}        \fi
\providecommand\bibfield[2]{#2}
\providecommand\bibinfo[2]{#2}
\providecommand\natexlab[1]{#1}
\providecommand\showeprint[2][]{arXiv:#2}

\bibitem[\protect\citeauthoryear{??}{ios}{2011}]%
        {ios}
 \bibinfo{year}{2011}\natexlab{}.
\newblock \bibinfo{title}{iOS - Siri - Apple}.
\newblock
\newblock
\urldef\tempurl%
\url{https://www.apple.com/ios/siri/}
\showURL{%
\tempurl}


\bibitem[\protect\citeauthoryear{??}{ale}{2014}]%
        {alexa}
 \bibinfo{year}{2014}\natexlab{}.
\newblock \bibinfo{title}{Alexa - Amazon}.
\newblock
\newblock
\urldef\tempurl%
\url{https://developer.amazon.com/alexa}
\showURL{%
\tempurl}


\bibitem[\protect\citeauthoryear{??}{goo}{2014}]%
        {google_cloud}
 \bibinfo{year}{2014}\natexlab{}.
\newblock \bibinfo{title}{Google Cloud AI}.
\newblock
\newblock
\urldef\tempurl%
\url{https://cloud.google.com/products/machine-learning/}
\showURL{%
\tempurl}


\bibitem[\protect\citeauthoryear{??}{cor}{2015}]%
        {cortana}
 \bibinfo{year}{2015}\natexlab{}.
\newblock \bibinfo{title}{Cortana | Your Intelligent Virtual \& Personal
  Assistant | Microsoft}.
\newblock
\newblock
\urldef\tempurl%
\url{https://www.microsoft.com/en-us/windows/cortana}
\showURL{%
\tempurl}


\bibitem[\protect\citeauthoryear{??}{nvi}{2015}]%
        {nvidia_cloud}
 \bibinfo{year}{2015}\natexlab{}.
\newblock \bibinfo{title}{nVIDIA GPU Cloud Computing}.
\newblock
\newblock
\urldef\tempurl%
\url{http://www.nvidia.com/object/gpu-cloud-computing.html}
\showURL{%
\tempurl}


\bibitem[\protect\citeauthoryear{??}{ali}{2017}]%
        {alibaba_cloud}
 \bibinfo{year}{2017}\natexlab{}.
\newblock \bibinfo{title}{Alibaba Cloud}.
\newblock
\newblock
\urldef\tempurl%
\url{https://www.alibabacloud.com/}
\showURL{%
\tempurl}


\bibitem[\protect\citeauthoryear{??}{int}{2019}]%
        {intel_cloud}
 \bibinfo{year}{2019}\natexlab{}.
\newblock \bibinfo{title}{Intel Nervana Platform}.
\newblock
\newblock
\urldef\tempurl%
\url{https://www.intelnervana.com/intel-nervana-platform/}
\showURL{%
\tempurl}


\bibitem[\protect\citeauthoryear{Abadi, Chu, Goodfellow, McMahan, Mironov,
  Talwar, and Zhang}{Abadi et~al\mbox{.}}{2016}]%
        {abadi2016deep}
\bibfield{author}{\bibinfo{person}{Mart{\'\i}n Abadi}, \bibinfo{person}{Andy
  Chu}, \bibinfo{person}{Ian Goodfellow}, \bibinfo{person}{H~Brendan McMahan},
  \bibinfo{person}{Ilya Mironov}, \bibinfo{person}{Kunal Talwar}, {and}
  \bibinfo{person}{Li Zhang}.} \bibinfo{year}{2016}\natexlab{}.
\newblock \showarticletitle{Deep learning with differential privacy}. In
  \bibinfo{booktitle}{\emph{Proceedings of the 2016 ACM SIGSAC Conference on
  Computer and Communications Security}}. ACM, \bibinfo{pages}{308--318}.
\newblock


\bibitem[\protect\citeauthoryear{Akhtar and Mian}{Akhtar and Mian}{2018}]%
        {akhtar2018threat}
\bibfield{author}{\bibinfo{person}{Naveed Akhtar} {and} \bibinfo{person}{Ajmal
  Mian}.} \bibinfo{year}{2018}\natexlab{}.
\newblock \showarticletitle{Threat of Adversarial Attacks on Deep Learning in
  Computer Vision: A Survey}.
\newblock \bibinfo{journal}{\emph{arXiv preprint arXiv:1801.00553}}
  (\bibinfo{year}{2018}).
\newblock


\bibitem[\protect\citeauthoryear{Alec~Radford and Chintala}{Alec~Radford and
  Chintala}{2018}]%
        {radford2018dcgan}
\bibfield{author}{\bibinfo{person}{Luke~Metz Alec~Radford} {and}
  \bibinfo{person}{Soumith Chintala}.} \bibinfo{year}{2018}\natexlab{}.
\newblock \showarticletitle{Unsupervised Representation Learning with Deep
  Convolutional Generative Adversarial Networks}.
\newblock \bibinfo{journal}{\emph{arXiv preprint arXiv:1511.06434}}
  (\bibinfo{year}{2018}).
\newblock


\bibitem[\protect\citeauthoryear{Ateniese, Felici, Mancini, Spognardi, Villani,
  and Vitali}{Ateniese et~al\mbox{.}}{2013}]%
        {AtenieseFMSVV13}
\bibfield{author}{\bibinfo{person}{Giuseppe Ateniese},
  \bibinfo{person}{Giovanni Felici}, \bibinfo{person}{Luigi~V. Mancini},
  \bibinfo{person}{Angelo Spognardi}, \bibinfo{person}{Antonio Villani}, {and}
  \bibinfo{person}{Domenico Vitali}.} \bibinfo{year}{2013}\natexlab{}.
\newblock \showarticletitle{Hacking Smart Machines with Smarter Ones: How to
  Extract Meaningful Data from Machine Learning Classifiers}.
\newblock \bibinfo{journal}{\emph{CoRR}}  \bibinfo{volume}{abs/1306.4447}
  (\bibinfo{year}{2013}).
\newblock
\showeprint[arxiv]{1306.4447}
\urldef\tempurl%
\url{http://arxiv.org/abs/1306.4447}
\showURL{%
\tempurl}


\bibitem[\protect\citeauthoryear{Barreno, Nelson, Joseph, and Tygar}{Barreno
  et~al\mbox{.}}{2010}]%
        {barreno2010security}
\bibfield{author}{\bibinfo{person}{Marco Barreno}, \bibinfo{person}{Blaine
  Nelson}, \bibinfo{person}{Anthony~D Joseph}, {and} \bibinfo{person}{JD
  Tygar}.} \bibinfo{year}{2010}\natexlab{}.
\newblock \showarticletitle{The security of machine learning}.
\newblock \bibinfo{journal}{\emph{Machine Learning}} \bibinfo{volume}{81},
  \bibinfo{number}{2} (\bibinfo{year}{2010}), \bibinfo{pages}{121--148}.
\newblock


\bibitem[\protect\citeauthoryear{Barreno, Nelson, Sears, Joseph, and
  Tygar}{Barreno et~al\mbox{.}}{2006}]%
        {barreno2006can}
\bibfield{author}{\bibinfo{person}{Marco Barreno}, \bibinfo{person}{Blaine
  Nelson}, \bibinfo{person}{Russell Sears}, \bibinfo{person}{Anthony~D Joseph},
  {and} \bibinfo{person}{J~Doug Tygar}.} \bibinfo{year}{2006}\natexlab{}.
\newblock \showarticletitle{Can machine learning be secure?}. In
  \bibinfo{booktitle}{\emph{Proceedings of the 2006 ACM Symposium on
  Information, computer and communications security}}. ACM,
  \bibinfo{pages}{16--25}.
\newblock


\bibitem[\protect\citeauthoryear{Benjamin I. P.~Rubinstein}{Benjamin I.
  P.~Rubinstein}{2009}]%
        {rubinstein2009privacysvm}
\bibfield{author}{\bibinfo{person}{Ling Huang Nina~Taft Benjamin I.
  P.~Rubinstein, Peter L.~Bartlett}.} \bibinfo{year}{2009}\natexlab{}.
\newblock \showarticletitle{Learning in a Large Function Space:
  Privacy-Preserving Mechanisms for SVM Learning}.
\newblock \bibinfo{journal}{\emph{arXiv preprint arXiv:0911.5708, 2009}}
  (\bibinfo{year}{2009}).
\newblock


\bibitem[\protect\citeauthoryear{Biggio, Corona, Nelson, Rubinstein, Maiorca,
  Fumera, Giacinto, and Roli}{Biggio et~al\mbox{.}}{2014a}]%
        {biggio2014svm}
\bibfield{author}{\bibinfo{person}{Battista Biggio}, \bibinfo{person}{Igino
  Corona}, \bibinfo{person}{Blaine Nelson}, \bibinfo{person}{Benjamin I.~P.
  Rubinstein}, \bibinfo{person}{Davide Maiorca}, \bibinfo{person}{Giorgio
  Fumera}, \bibinfo{person}{Giorgio Giacinto}, {and} \bibinfo{person}{Fabio
  Roli}.} \bibinfo{year}{2014}\natexlab{a}.
\newblock \bibinfo{booktitle}{\emph{Security Evaluation of Support Vector
  Machines in Adversarial Environments}}.
\newblock \bibinfo{publisher}{Springer International Publishing},
  \bibinfo{address}{Cham}, \bibinfo{pages}{105--153}.
\newblock
\urldef\tempurl%
\url{https://doi.org/10.1007/978-3-319-02300-7_4}
\showDOI{\tempurl}


\bibitem[\protect\citeauthoryear{Biggio, Fumera, and Roli}{Biggio
  et~al\mbox{.}}{2008}]%
        {biggio2008adversarial}
\bibfield{author}{\bibinfo{person}{Battista Biggio}, \bibinfo{person}{Giorgio
  Fumera}, {and} \bibinfo{person}{Fabio Roli}.}
  \bibinfo{year}{2008}\natexlab{}.
\newblock \showarticletitle{Adversarial pattern classification using multiple
  classifiers and randomisation}.
\newblock \bibinfo{journal}{\emph{Structural, Syntactic, and Statistical
  Pattern Recognition}} (\bibinfo{year}{2008}), \bibinfo{pages}{500--509}.
\newblock


\bibitem[\protect\citeauthoryear{Biggio, Fumera, and Roli}{Biggio
  et~al\mbox{.}}{2014b}]%
        {biggio2014securityevaluation}
\bibfield{author}{\bibinfo{person}{Battista Biggio}, \bibinfo{person}{Giorgio
  Fumera}, {and} \bibinfo{person}{Fabio Roli}.}
  \bibinfo{year}{2014}\natexlab{b}.
\newblock \showarticletitle{Security Evaluation of Pattern Classifiers under
  Attack}.
\newblock \bibinfo{journal}{\emph{{IEEE} Trans. Knowl. Data Eng.}}
  \bibinfo{volume}{26}, \bibinfo{number}{4} (\bibinfo{year}{2014}),
  \bibinfo{pages}{984--996}.
\newblock
\urldef\tempurl%
\url{https://doi.org/10.1109/TKDE.2013.57}
\showDOI{\tempurl}


\bibitem[\protect\citeauthoryear{Biggio, Fumera, and Roli}{Biggio
  et~al\mbox{.}}{2014c}]%
        {biggio2014security}
\bibfield{author}{\bibinfo{person}{Battista Biggio}, \bibinfo{person}{Giorgio
  Fumera}, {and} \bibinfo{person}{Fabio Roli}.}
  \bibinfo{year}{2014}\natexlab{c}.
\newblock \showarticletitle{Security evaluation of pattern classifiers under
  attack}.
\newblock \bibinfo{journal}{\emph{IEEE transactions on knowledge and data
  engineering}} \bibinfo{volume}{26}, \bibinfo{number}{4}
  (\bibinfo{year}{2014}), \bibinfo{pages}{984--`--996}.
\newblock


\bibitem[\protect\citeauthoryear{Biggio, Nelson, and Laskov}{Biggio
  et~al\mbox{.}}{2011}]%
        {Biggio2011support}
\bibfield{author}{\bibinfo{person}{Battista Biggio}, \bibinfo{person}{Blaine
  Nelson}, {and} \bibinfo{person}{Pavel Laskov}.}
  \bibinfo{year}{2011}\natexlab{}.
\newblock \showarticletitle{Support Vector Machines Under Adversarial Label
  Noise}. In \bibinfo{booktitle}{\emph{Proceedings of the 3rd Asian Conference
  on Machine Learning, {ACML} 2011, Taoyuan, Taiwan, November 13-15, 2011}}.
  \bibinfo{pages}{97--112}.
\newblock
\urldef\tempurl%
\url{http://www.jmlr.org/proceedings/papers/v20/biggio11/biggio11.pdf}
\showURL{%
\tempurl}


\bibitem[\protect\citeauthoryear{Biggio, Nelson, and Laskov}{Biggio
  et~al\mbox{.}}{2012}]%
        {Biggio2012poisoning}
\bibfield{author}{\bibinfo{person}{Battista Biggio}, \bibinfo{person}{Blaine
  Nelson}, {and} \bibinfo{person}{Pavel Laskov}.}
  \bibinfo{year}{2012}\natexlab{}.
\newblock \showarticletitle{Poisoning Attacks against Support Vector Machines}.
  In \bibinfo{booktitle}{\emph{Proceedings of the 29th International Conference
  on Machine Learning, {ICML} 2012, Edinburgh, Scotland, UK, June 26 - July 1,
  2012}}.
\newblock


\bibitem[\protect\citeauthoryear{Biggio and Roli}{Biggio and Roli}{2014}]%
        {biggio2014secevaluation}
\bibfield{author}{\bibinfo{person}{Corona I. Nelson B. Rubinstein B. I. Maiorca
  D. Fumera G. Giacinto~G. Biggio, B.} {and} \bibinfo{person}{F Roli}.}
  \bibinfo{year}{2014}\natexlab{}.
\newblock \showarticletitle{Security evaluation of support vector machines in
  adversarial environments}.
\newblock  (\bibinfo{year}{2014}).
\newblock


\bibitem[\protect\citeauthoryear{Carlini and Wagner}{Carlini and
  Wagner}{2017}]%
        {carlini2017towards}
\bibfield{author}{\bibinfo{person}{Nicholas Carlini} {and}
  \bibinfo{person}{David Wagner}.} \bibinfo{year}{2017}\natexlab{}.
\newblock \showarticletitle{Towards evaluating the robustness of neural
  networks}. In \bibinfo{booktitle}{\emph{Security and Privacy (SP), 2017 IEEE
  Symposium on}}. IEEE, \bibinfo{pages}{39--57}.
\newblock


\bibitem[\protect\citeauthoryear{Corona, Giacinto, and Roli}{Corona
  et~al\mbox{.}}{2013}]%
        {corona2013adversarial}
\bibfield{author}{\bibinfo{person}{Igino Corona}, \bibinfo{person}{Giorgio
  Giacinto}, {and} \bibinfo{person}{Fabio Roli}.}
  \bibinfo{year}{2013}\natexlab{}.
\newblock \showarticletitle{Adversarial attacks against intrusion detection
  systems: Taxonomy, solutions and open issues}.
\newblock \bibinfo{journal}{\emph{Information Sciences}}  \bibinfo{volume}{239}
  (\bibinfo{year}{2013}), \bibinfo{pages}{201--225}.
\newblock


\bibitem[\protect\citeauthoryear{Dalvi, Domingos, Sanghai, Verma,
  et~al\mbox{.}}{Dalvi et~al\mbox{.}}{2004}]%
        {dalvi2004adversarial}
\bibfield{author}{\bibinfo{person}{Nilesh Dalvi}, \bibinfo{person}{Pedro
  Domingos}, \bibinfo{person}{Sumit Sanghai}, \bibinfo{person}{Deepak Verma},
  {et~al\mbox{.}}} \bibinfo{year}{2004}\natexlab{}.
\newblock \showarticletitle{Adversarial classification}. In
  \bibinfo{booktitle}{\emph{Proceedings of the tenth ACM SIGKDD international
  conference on Knowledge discovery and data mining}}. ACM,
  \bibinfo{pages}{99--108}.
\newblock


\bibitem[\protect\citeauthoryear{Dongyu~Meng}{Dongyu~Meng}{2017}]%
        {meng2017magnet}
\bibfield{author}{\bibinfo{person}{Hao~Chen Dongyu~Meng}.}
  \bibinfo{year}{2017}\natexlab{}.
\newblock \showarticletitle{MagNet: a Two-Pronged Defense against Adversarial
  Examples}.
\newblock  (\bibinfo{year}{2017}).
\newblock


\bibitem[\protect\citeauthoryear{Fredrikson, Jha, and Ristenpart}{Fredrikson
  et~al\mbox{.}}{2015}]%
        {fredrikson2015model}
\bibfield{author}{\bibinfo{person}{Matt Fredrikson}, \bibinfo{person}{Somesh
  Jha}, {and} \bibinfo{person}{Thomas Ristenpart}.}
  \bibinfo{year}{2015}\natexlab{}.
\newblock \showarticletitle{Model inversion attacks that exploit confidence
  information and basic countermeasures}. In
  \bibinfo{booktitle}{\emph{Proceedings of the 22nd ACM SIGSAC Conference on
  Computer and Communications Security}}. ACM, \bibinfo{pages}{1322--1333}.
\newblock


\bibitem[\protect\citeauthoryear{Fredrikson, Lantz, Jha, Lin, Page, and
  Ristenpart}{Fredrikson et~al\mbox{.}}{[n. d.]}]%
        {fredrikson2014privacy}
\bibfield{author}{\bibinfo{person}{Matthew Fredrikson}, \bibinfo{person}{Eric
  Lantz}, \bibinfo{person}{Somesh Jha}, \bibinfo{person}{Simon Lin},
  \bibinfo{person}{David Page}, {and} \bibinfo{person}{Thomas Ristenpart}.}
  \bibinfo{year}{[n. d.]}\natexlab{}.
\newblock \showarticletitle{Privacy in Pharmacogenetics: An End-to-End Case
  Study of Personalized Warfarin Dosing.}
\newblock


\bibitem[\protect\citeauthoryear{G.~Zhang and Xu}{G.~Zhang and Xu}{2017}]%
        {zhang2017vcs}
\bibfield{author}{\bibinfo{person}{X.~Ji T. Zhang T.~Zhang G.~Zhang, C.~Yan}
  {and} \bibinfo{person}{W. Xu}.} \bibinfo{year}{2017}\natexlab{}.
\newblock \showarticletitle{Dolphinatack: Inaudible voice commands}.
\newblock \bibinfo{journal}{\emph{arXiv preprint arXiv:1708.09537}}
  (\bibinfo{year}{2017}).
\newblock


\bibitem[\protect\citeauthoryear{Golland}{Golland}{2001}]%
        {golland2001discriminative}
\bibfield{author}{\bibinfo{person}{Polina Golland}.}
  \bibinfo{year}{2001}\natexlab{}.
\newblock \showarticletitle{Discriminative Direction for Kernel Classifiers}.
\newblock  (\bibinfo{year}{2001}).
\newblock


\bibitem[\protect\citeauthoryear{Goodfellow, Pouget{-}Abadie, Mirza, Xu,
  Warde{-}Farley, Ozair, Courville, and Bengio}{Goodfellow
  et~al\mbox{.}}{2014a}]%
        {goodfellow2014gan}
\bibfield{author}{\bibinfo{person}{Ian~J. Goodfellow}, \bibinfo{person}{Jean
  Pouget{-}Abadie}, \bibinfo{person}{Mehdi Mirza}, \bibinfo{person}{Bing Xu},
  \bibinfo{person}{David Warde{-}Farley}, \bibinfo{person}{Sherjil Ozair},
  \bibinfo{person}{Aaron~C. Courville}, {and} \bibinfo{person}{Yoshua Bengio}.}
  \bibinfo{year}{2014}\natexlab{a}.
\newblock \showarticletitle{Generative Adversarial Networks}.
\newblock \bibinfo{journal}{\emph{CoRR}}  \bibinfo{volume}{abs/1406.2661}
  (\bibinfo{year}{2014}).
\newblock
\showeprint[arxiv]{1406.2661}
\urldef\tempurl%
\url{http://arxiv.org/abs/1406.2661}
\showURL{%
\tempurl}


\bibitem[\protect\citeauthoryear{Goodfellow, Shlens, and Szegedy}{Goodfellow
  et~al\mbox{.}}{2014b}]%
        {goodfellow2014explaining}
\bibfield{author}{\bibinfo{person}{Ian~J. Goodfellow},
  \bibinfo{person}{Jonathon Shlens}, {and} \bibinfo{person}{Christian
  Szegedy}.} \bibinfo{year}{2014}\natexlab{b}.
\newblock \showarticletitle{Explaining and Harnessing Adversarial Examples}.
\newblock \bibinfo{journal}{\emph{CoRR}}  \bibinfo{volume}{abs/1412.6572}
  (\bibinfo{year}{2014}).
\newblock
\urldef\tempurl%
\url{http://arxiv.org/abs/1412.6572}
\showURL{%
\tempurl}


\bibitem[\protect\citeauthoryear{H.~Y.~Xiong and Morris}{H.~Y.~Xiong and
  Morris}{2015}]%
        {xiong2015gene}
\bibfield{author}{\bibinfo{person}{J.~L. Lee H. Bretschneider D. Merico
  R.K.~Yuen H.~Y.~Xiong, B.~Alipanahi} {and} \bibinfo{person}{Q. Morris}.}
  \bibinfo{year}{2015}\natexlab{}.
\newblock \showarticletitle{The human splicing code reveals new insights into
  the genetic determinants of disease}.
\newblock   \bibinfo{volume}{347, no. 6218} (\bibinfo{year}{2015}).
\newblock


\bibitem[\protect\citeauthoryear{Hinton, Vinyals, and Dean}{Hinton
  et~al\mbox{.}}{2015}]%
        {hinton2015distillating}
\bibfield{author}{\bibinfo{person}{Geoffrey~E. Hinton}, \bibinfo{person}{Oriol
  Vinyals}, {and} \bibinfo{person}{Jeffrey Dean}.}
  \bibinfo{year}{2015}\natexlab{}.
\newblock \showarticletitle{Distilling the Knowledge in a Neural Network}.
\newblock \bibinfo{journal}{\emph{CoRR}}  \bibinfo{volume}{abs/1503.02531}
  (\bibinfo{year}{2015}).
\newblock
\showeprint[arxiv]{1503.02531}
\urldef\tempurl%
\url{http://arxiv.org/abs/1503.02531}
\showURL{%
\tempurl}


\bibitem[\protect\citeauthoryear{Hitaj, Ateniese, and Perez-Cruz}{Hitaj
  et~al\mbox{.}}{2017}]%
        {hitaj2017deep}
\bibfield{author}{\bibinfo{person}{Briland Hitaj}, \bibinfo{person}{Giuseppe
  Ateniese}, {and} \bibinfo{person}{Fernando Perez-Cruz}.}
  \bibinfo{year}{2017}\natexlab{}.
\newblock \showarticletitle{Deep Models Under the GAN: Information Leakage from
  Collaborative Deep Learning}.
\newblock \bibinfo{journal}{\emph{arXiv preprint arXiv:1702.07464}}
  (\bibinfo{year}{2017}).
\newblock


\bibitem[\protect\citeauthoryear{Hosseini, Chen, Kannan, Zhang, and
  Poovendran}{Hosseini et~al\mbox{.}}{2017}]%
        {hossain2017blocking}
\bibfield{author}{\bibinfo{person}{Hossein Hosseini}, \bibinfo{person}{Yize
  Chen}, \bibinfo{person}{Sreeram Kannan}, \bibinfo{person}{Baosen Zhang},
  {and} \bibinfo{person}{Radha Poovendran}.} \bibinfo{year}{2017}\natexlab{}.
\newblock \showarticletitle{Blocking Transferability of Adversarial Examples in
  Black-Box Learning Systems}.
\newblock \bibinfo{journal}{\emph{CoRR}}  \bibinfo{volume}{abs/1703.04318}
  (\bibinfo{year}{2017}).
\newblock
\showeprint[arxiv]{1703.04318}
\urldef\tempurl%
\url{http://arxiv.org/abs/1703.04318}
\showURL{%
\tempurl}


\bibitem[\protect\citeauthoryear{Ioffe and Szegedy}{Ioffe and Szegedy}{2015}]%
        {sergey2015batchnorm}
\bibfield{author}{\bibinfo{person}{Sergey Ioffe} {and}
  \bibinfo{person}{Christian Szegedy}.} \bibinfo{year}{2015}\natexlab{}.
\newblock \showarticletitle{Batch normalization: Accelerating deep network
  training by reducing internal covariate shift}.
\newblock \bibinfo{journal}{\emph{arXiv preprint arXiv:1502.03167, 2015}}
  (\bibinfo{year}{2015}).
\newblock


\bibitem[\protect\citeauthoryear{J.~Ma and Svetnik}{J.~Ma and Svetnik}{2015}]%
        {ma2015structure}
\bibfield{author}{\bibinfo{person}{A.~Liaw G. E.~Dahl J.~Ma, R. P.~Sheridan}
  {and} \bibinfo{person}{V. Svetnik}.} \bibinfo{year}{2015}\natexlab{}.
\newblock \showarticletitle{Deep neural nets as a method for quantitative
  structure-activity relationships}.
\newblock \bibinfo{journal}{\emph{Journal of chemical information and
  modeling}}  \bibinfo{volume}{55, no. 2} (\bibinfo{year}{2015}),
  \bibinfo{pages}{263--274}.
\newblock


\bibitem[\protect\citeauthoryear{Jian-Feng~Cai and Shen}{Jian-Feng~Cai and
  Shen}{2010}]%
        {cai2010matcomp}
\bibfield{author}{\bibinfo{person}{Emmanuel J.~Candès Jian-Feng~Cai} {and}
  \bibinfo{person}{Zuowei Shen}.} \bibinfo{year}{2010}\natexlab{}.
\newblock \showarticletitle{A singular value thresholding algorithm for matrix
  completion}.
\newblock   \bibinfo{volume}{20} (\bibinfo{year}{2010}), \bibinfo{pages}{1956
  -- 1982}.
\newblock
Issue 4.


\bibitem[\protect\citeauthoryear{Kaggle}{Kaggle}{2014}]%
        {kaggle2012higgs}
\bibfield{author}{\bibinfo{person}{Kaggle}.} \bibinfo{year}{2014}\natexlab{}.
\newblock \showarticletitle{Higgs boson machine learning challenge}.
\newblock  (\bibinfo{year}{2014}).
\newblock
\urldef\tempurl%
\url{https://www.kaggle.com/c/higgs-boson, 2014}
\showURL{%
\tempurl}


\bibitem[\protect\citeauthoryear{Kearns and Li}{Kearns and Li}{1993}]%
        {Kearns1993learning}
\bibfield{author}{\bibinfo{person}{Michael Kearns} {and} \bibinfo{person}{Ming
  Li}.} \bibinfo{year}{1993}\natexlab{}.
\newblock \showarticletitle{Learning in the Presence of Malicious Errors}.
\newblock \bibinfo{journal}{\emph{SIAM J. Comput.}} \bibinfo{volume}{22},
  \bibinfo{number}{4} (\bibinfo{date}{aug} \bibinfo{year}{1993}),
  \bibinfo{pages}{807--837}.
\newblock
\showISSN{0097-5397}
\urldef\tempurl%
\url{https://doi.org/10.1137/0222052}
\showDOI{\tempurl}


\bibitem[\protect\citeauthoryear{Kloft and Laskov}{Kloft and Laskov}{2010}]%
        {kloft2010online}
\bibfield{author}{\bibinfo{person}{Marius Kloft} {and} \bibinfo{person}{Pavel
  Laskov}.} \bibinfo{year}{2010}\natexlab{}.
\newblock \showarticletitle{Online Anomaly Detection under Adversarial Impact}.
  In \bibinfo{booktitle}{\emph{Proceedings of the Thirteenth International
  Conference on Artificial Intelligence and Statistics}}
  \emph{(\bibinfo{series}{Proceedings of Machine Learning Research})},
  \bibfield{editor}{\bibinfo{person}{Yee~Whye Teh} {and} \bibinfo{person}{Mike
  Titterington}} (Eds.), Vol.~\bibinfo{volume}{9}. \bibinfo{publisher}{PMLR},
  \bibinfo{address}{Chia Laguna Resort, Sardinia, Italy},
  \bibinfo{pages}{405--412}.
\newblock
\urldef\tempurl%
\url{http://proceedings.mlr.press/v9/kloft10a.html}
\showURL{%
\tempurl}


\bibitem[\protect\citeauthoryear{Kumar and Mehta}{Kumar and Mehta}{2017}]%
        {DBLP:journals/corr/KumarM17}
\bibfield{author}{\bibinfo{person}{Atul Kumar} {and} \bibinfo{person}{Sameep
  Mehta}.} \bibinfo{year}{2017}\natexlab{}.
\newblock \showarticletitle{A Survey on Resilient Machine Learning}.
\newblock \bibinfo{journal}{\emph{CoRR}}  \bibinfo{volume}{abs/1707.03184}
  (\bibinfo{year}{2017}).
\newblock
\showeprint[arxiv]{1707.03184}
\urldef\tempurl%
\url{http://arxiv.org/abs/1707.03184}
\showURL{%
\tempurl}


\bibitem[\protect\citeauthoryear{Kurakin, Goodfellow, and Bengio}{Kurakin
  et~al\mbox{.}}{2016}]%
        {kurakin2016adversarial}
\bibfield{author}{\bibinfo{person}{Alexey Kurakin}, \bibinfo{person}{Ian~J.
  Goodfellow}, {and} \bibinfo{person}{Samy Bengio}.}
  \bibinfo{year}{2016}\natexlab{}.
\newblock \showarticletitle{Adversarial Machine Learning at Scale}.
\newblock \bibinfo{journal}{\emph{CoRR}}  \bibinfo{volume}{abs/1611.01236}
  (\bibinfo{year}{2016}).
\newblock
\showeprint[arxiv]{1611.01236}
\urldef\tempurl%
\url{http://arxiv.org/abs/1611.01236}
\showURL{%
\tempurl}


\bibitem[\protect\citeauthoryear{LeCun, Bengio, and Hinton}{LeCun
  et~al\mbox{.}}{2015}]%
        {lecun2015deep}
\bibfield{author}{\bibinfo{person}{Yann LeCun}, \bibinfo{person}{Yoshua
  Bengio}, {and} \bibinfo{person}{Geoffrey Hinton}.}
  \bibinfo{year}{2015}\natexlab{}.
\newblock \showarticletitle{Deep learning}.
\newblock \bibinfo{journal}{\emph{nature}} \bibinfo{volume}{521},
  \bibinfo{number}{7553} (\bibinfo{year}{2015}), \bibinfo{pages}{436}.
\newblock


\bibitem[\protect\citeauthoryear{Li, Wang, Singh, and Vorobeychik}{Li
  et~al\mbox{.}}{2016}]%
        {Li2016data}
\bibfield{author}{\bibinfo{person}{Bo Li}, \bibinfo{person}{Yining Wang},
  \bibinfo{person}{Aarti Singh}, {and} \bibinfo{person}{Yevgeniy Vorobeychik}.}
  \bibinfo{year}{2016}\natexlab{}.
\newblock \showarticletitle{Data Poisoning Attacks on Factorization-Based
  Collaborative Filtering}.
\newblock \bibinfo{journal}{\emph{CoRR}}  \bibinfo{volume}{abs/1608.08182}
  (\bibinfo{year}{2016}).
\newblock
\showeprint[arxiv]{1608.08182}
\urldef\tempurl%
\url{http://arxiv.org/abs/1608.08182}
\showURL{%
\tempurl}


\bibitem[\protect\citeauthoryear{Liao, Liang, Dong, Pang, Zhu, and Hu}{Liao
  et~al\mbox{.}}{2017}]%
        {liao2017hgd}
\bibfield{author}{\bibinfo{person}{Fangzhou Liao}, \bibinfo{person}{Ming
  Liang}, \bibinfo{person}{Yinpeng Dong}, \bibinfo{person}{Tianyu Pang},
  \bibinfo{person}{Jun Zhu}, {and} \bibinfo{person}{Xiaolin Hu}.}
  \bibinfo{year}{2017}\natexlab{}.
\newblock \showarticletitle{Defense against Adversarial Attacks Using
  High-Level Representation Guided Denoiser}.
\newblock \bibinfo{journal}{\emph{arXiv preprint arXiv:1705.09064}}
  (\bibinfo{year}{2017}).
\newblock


\bibitem[\protect\citeauthoryear{Liu and Nocedal}{Liu and Nocedal}{1989}]%
        {liu1989limited}
\bibfield{author}{\bibinfo{person}{Dong~C Liu} {and} \bibinfo{person}{Jorge
  Nocedal}.} \bibinfo{year}{1989}\natexlab{}.
\newblock \showarticletitle{On the limited memory BFGS method for large scale
  optimization}.
\newblock \bibinfo{journal}{\emph{Mathematical programming}}
  \bibinfo{volume}{45}, \bibinfo{number}{1} (\bibinfo{year}{1989}),
  \bibinfo{pages}{503--528}.
\newblock


\bibitem[\protect\citeauthoryear{Lyu, Huang, and Liang}{Lyu
  et~al\mbox{.}}{2015}]%
        {lyu2015aunified}
\bibfield{author}{\bibinfo{person}{Chunchuan Lyu}, \bibinfo{person}{Kaizhu
  Huang}, {and} \bibinfo{person}{Hai{-}Ning Liang}.}
  \bibinfo{year}{2015}\natexlab{}.
\newblock \showarticletitle{A Unified Gradient Regularization Family for
  Adversarial Examples}. In \bibinfo{booktitle}{\emph{2015 {IEEE} International
  Conference on Data Mining, {ICDM} 2015, Atlantic City, NJ, USA, November
  14-17, 2015}}. \bibinfo{pages}{301--309}.
\newblock
\urldef\tempurl%
\url{https://doi.org/10.1109/ICDM.2015.84}
\showDOI{\tempurl}


\bibitem[\protect\citeauthoryear{M.~Helmstaedter and Denk}{M.~Helmstaedter and
  Denk}{2013}]%
        {helmstaedter2013retina}
\bibfield{author}{\bibinfo{person}{S.~C. Turaga V. Jain H. S.~Seung
  M.~Helmstaedter, K. L.~Briggman} {and} \bibinfo{person}{W. Denk}.}
  \bibinfo{year}{2013}\natexlab{}.
\newblock \showarticletitle{Connectomic reconstruction of the inner plexiform
  layer in the mouse retina}.
\newblock   \bibinfo{volume}{500, no. 7461} (\bibinfo{year}{2013}),
  \bibinfo{pages}{168--147}.
\newblock


\bibitem[\protect\citeauthoryear{Maas and Ng}{Maas and Ng}{2013}]%
        {mass2013leakyrelu}
\bibfield{author}{\bibinfo{person}{Hannun Awni~Y Maas, Andrew~L} {and}
  \bibinfo{person}{Andrew~Y Ng}.} \bibinfo{year}{2013}\natexlab{}.
\newblock \showarticletitle{Rectifier nonlinearities improve neural network
  acoustic models}.
\newblock   \bibinfo{volume}{30} (\bibinfo{year}{2013}).
\newblock


\bibitem[\protect\citeauthoryear{Mei and Zhu}{Mei and Zhu}{2015}]%
        {Mei2015using}
\bibfield{author}{\bibinfo{person}{Shike Mei} {and} \bibinfo{person}{Xiaojin
  Zhu}.} \bibinfo{year}{2015}\natexlab{}.
\newblock \showarticletitle{Using Machine Teaching to Identify Optimal
  Training-set Attacks on Machine Learners}. In
  \bibinfo{booktitle}{\emph{Proceedings of the Twenty-Ninth AAAI Conference on
  Artificial Intelligence}} \emph{(\bibinfo{series}{AAAI'15})}.
  \bibinfo{publisher}{AAAI Press}, \bibinfo{pages}{2871--2877}.
\newblock
\showISBNx{0-262-51129-0}
\urldef\tempurl%
\url{http://dl.acm.org/citation.cfm?id=2886521.2886721}
\showURL{%
\tempurl}


\bibitem[\protect\citeauthoryear{Moosavi-Dezfooli, Fawzi, and
  Frossard}{Moosavi-Dezfooli et~al\mbox{.}}{2016}]%
        {moosavi2016deepfool}
\bibfield{author}{\bibinfo{person}{Seyed-Mohsen Moosavi-Dezfooli},
  \bibinfo{person}{Alhussein Fawzi}, {and} \bibinfo{person}{Pascal Frossard}.}
  \bibinfo{year}{2016}\natexlab{}.
\newblock \showarticletitle{Deepfool: a simple and accurate method to fool deep
  neural networks}. In \bibinfo{booktitle}{\emph{Proceedings of the IEEE
  Conference on Computer Vision and Pattern Recognition}}.
  \bibinfo{pages}{2574--2582}.
\newblock


\bibitem[\protect\citeauthoryear{Mordvintsev and Tyka}{Mordvintsev and
  Tyka}{2015}]%
        {mordvintsev2015inception}
\bibfield{author}{\bibinfo{person}{Olah-Christopher Mordvintsev, Alexander}
  {and} \bibinfo{person}{Mike Tyka}.} \bibinfo{year}{2015}\natexlab{}.
\newblock \showarticletitle{Inceptionism : Going deeper into neural networks}.
\newblock  (\bibinfo{year}{2015}).
\newblock
\urldef\tempurl%
\url{https://ai.googleblog.com/2015/06/inceptionism-going-deeper-into-neural.html}
\showURL{%
\tempurl}


\bibitem[\protect\citeauthoryear{N.~Carlini and Zhou}{N.~Carlini and
  Zhou}{2016}]%
        {carlini2016asr}
\bibfield{author}{\bibinfo{person}{T.~Vaidya Y. Zhang M. Sherr C. Shields
  D.~Wagner N.~Carlini, P.~Mishra} {and} \bibinfo{person}{W. Zhou}.}
  \bibinfo{year}{2016}\natexlab{}.
\newblock \showarticletitle{Hidden voice commands}.
\newblock  (\bibinfo{year}{2016}), \bibinfo{pages}{513--530}.
\newblock


\bibitem[\protect\citeauthoryear{Nair and Hinton}{Nair and Hinton}{2010}]%
        {nair2010relu}
\bibfield{author}{\bibinfo{person}{Vinod Nair} {and} \bibinfo{person}{Geoffrey
  Hinton}.} \bibinfo{year}{2010}\natexlab{}.
\newblock \showarticletitle{Rectified linear units improve restricted boltzmann
  machines}.
\newblock  (\bibinfo{year}{2010}), \bibinfo{pages}{807--814}.
\newblock


\bibitem[\protect\citeauthoryear{Narodytska and Kasiviswanathan}{Narodytska and
  Kasiviswanathan}{2017}]%
        {narodytska2017simple}
\bibfield{author}{\bibinfo{person}{Nina Narodytska} {and}
  \bibinfo{person}{Shiva~Prasad Kasiviswanathan}.}
  \bibinfo{year}{2017}\natexlab{}.
\newblock \showarticletitle{Simple Black-Box Adversarial Attacks on Deep Neural
  Networks}. In \bibinfo{booktitle}{\emph{2017 {IEEE} Conference on Computer
  Vision and Pattern Recognition Workshops, {CVPR} Workshops, Honolulu, HI,
  USA, July 21-26, 2017}}. \bibinfo{pages}{1310--1318}.
\newblock
\urldef\tempurl%
\url{https://doi.org/10.1109/CVPRW.2017.172}
\showDOI{\tempurl}


\bibitem[\protect\citeauthoryear{Papernot, McDaniel, and Goodfellow}{Papernot
  et~al\mbox{.}}{2016a}]%
        {papernot2016transferability}
\bibfield{author}{\bibinfo{person}{Nicolas Papernot}, \bibinfo{person}{Patrick
  McDaniel}, {and} \bibinfo{person}{Ian Goodfellow}.}
  \bibinfo{year}{2016}\natexlab{a}.
\newblock \showarticletitle{Transferability in machine learning: from phenomena
  to black-box attacks using adversarial samples}.
\newblock \bibinfo{journal}{\emph{arXiv preprint arXiv:1605.07277}}
  (\bibinfo{year}{2016}).
\newblock


\bibitem[\protect\citeauthoryear{Papernot, McDaniel, Jha, Fredrikson, Celik,
  and Swami}{Papernot et~al\mbox{.}}{2016b}]%
        {papernot2016limitations}
\bibfield{author}{\bibinfo{person}{Nicolas Papernot}, \bibinfo{person}{Patrick
  McDaniel}, \bibinfo{person}{Somesh Jha}, \bibinfo{person}{Matt Fredrikson},
  \bibinfo{person}{Z~Berkay Celik}, {and} \bibinfo{person}{Ananthram Swami}.}
  \bibinfo{year}{2016}\natexlab{b}.
\newblock \showarticletitle{The limitations of deep learning in adversarial
  settings}. In \bibinfo{booktitle}{\emph{Security and Privacy (EuroS\&P), 2016
  IEEE European Symposium on}}. IEEE, \bibinfo{pages}{372--387}.
\newblock


\bibitem[\protect\citeauthoryear{Papernot and McDaniel}{Papernot and
  McDaniel}{2017}]%
        {papernot2017extending}
\bibfield{author}{\bibinfo{person}{Nicolas Papernot} {and}
  \bibinfo{person}{Patrick~D. McDaniel}.} \bibinfo{year}{2017}\natexlab{}.
\newblock \showarticletitle{Extending Defensive Distillation}.
\newblock \bibinfo{journal}{\emph{CoRR}}  \bibinfo{volume}{abs/1705.05264}
  (\bibinfo{year}{2017}).
\newblock
\urldef\tempurl%
\url{http://arxiv.org/abs/1705.05264}
\showURL{%
\tempurl}


\bibitem[\protect\citeauthoryear{Papernot, McDaniel, Goodfellow, Jha, Celik,
  and Swami}{Papernot et~al\mbox{.}}{2017}]%
        {papernot2017practical}
\bibfield{author}{\bibinfo{person}{Nicolas Papernot},
  \bibinfo{person}{Patrick~D. McDaniel}, \bibinfo{person}{Ian~J. Goodfellow},
  \bibinfo{person}{Somesh Jha}, \bibinfo{person}{Z.~Berkay Celik}, {and}
  \bibinfo{person}{Ananthram Swami}.} \bibinfo{year}{2017}\natexlab{}.
\newblock \showarticletitle{Practical Black-Box Attacks against Machine
  Learning}. In \bibinfo{booktitle}{\emph{Proceedings of the 2017 {ACM} on Asia
  Conference on Computer and Communications Security, AsiaCCS 2017, Abu Dhabi,
  United Arab Emirates, April 2-6, 2017}}. \bibinfo{pages}{506--519}.
\newblock
\urldef\tempurl%
\url{https://doi.org/10.1145/3052973.3053009}
\showDOI{\tempurl}


\bibitem[\protect\citeauthoryear{Papernot, McDaniel, Sinha, and
  Wellman}{Papernot et~al\mbox{.}}{2016c}]%
        {papernot2016towards}
\bibfield{author}{\bibinfo{person}{Nicolas Papernot},
  \bibinfo{person}{Patrick~D. McDaniel}, \bibinfo{person}{Arunesh Sinha}, {and}
  \bibinfo{person}{Michael~P. Wellman}.} \bibinfo{year}{2016}\natexlab{c}.
\newblock \showarticletitle{Towards the Science of Security and Privacy in
  Machine Learning}.
\newblock \bibinfo{journal}{\emph{CoRR}}  \bibinfo{volume}{abs/1611.03814}
  (\bibinfo{year}{2016}).
\newblock
\urldef\tempurl%
\url{http://arxiv.org/abs/1611.03814}
\showURL{%
\tempurl}


\bibitem[\protect\citeauthoryear{Papernot, McDaniel, Wu, Jha, and
  Swami}{Papernot et~al\mbox{.}}{2016d}]%
        {papernot2016distillation}
\bibfield{author}{\bibinfo{person}{Nicolas Papernot},
  \bibinfo{person}{Patrick~D. McDaniel}, \bibinfo{person}{Xi Wu},
  \bibinfo{person}{Somesh Jha}, {and} \bibinfo{person}{Ananthram Swami}.}
  \bibinfo{year}{2016}\natexlab{d}.
\newblock \showarticletitle{Distillation as a Defense to Adversarial
  Perturbations Against Deep Neural Networks}. In
  \bibinfo{booktitle}{\emph{{IEEE} Symposium on Security and Privacy, {SP}
  2016, San Jose, CA, USA, May 22-26, 2016}}. \bibinfo{pages}{582--597}.
\newblock
\urldef\tempurl%
\url{https://doi.org/10.1109/SP.2016.41}
\showDOI{\tempurl}


\bibitem[\protect\citeauthoryear{Pouya~Samangouei and
  Chellappa}{Pouya~Samangouei and Chellappa}{2018}]%
        {samangouei2018defensegan}
\bibfield{author}{\bibinfo{person}{Maya~Kabkab Pouya~Samangouei} {and}
  \bibinfo{person}{Rama Chellappa}.} \bibinfo{year}{2018}\natexlab{}.
\newblock \showarticletitle{Defense-GAN: Protecting Classifiers Against
  Adversarial Attacks Using Generative Models}.
\newblock \bibinfo{journal}{\emph{arXiv preprint arXiv:1805.06605}}
  (\bibinfo{year}{2018}).
\newblock


\bibitem[\protect\citeauthoryear{Prateek~Jain and Sanghavi}{Prateek~Jain and
  Sanghavi}{2012}]%
        {jain2012altmin}
\bibfield{author}{\bibinfo{person}{Praneeth~Netrapalli Prateek~Jain} {and}
  \bibinfo{person}{Sujay Sanghavi}.} \bibinfo{year}{2012}\natexlab{}.
\newblock \showarticletitle{Low-rank matrix completion using alternating
  minimization}.
\newblock \bibinfo{journal}{\emph{arXiv preprint arXiv:1212.0467, 2012}}
  (\bibinfo{year}{2012}).
\newblock


\bibitem[\protect\citeauthoryear{Rosenberg, Shabtai, Rokach, and
  Elovici}{Rosenberg et~al\mbox{.}}{2017}]%
        {rosenberg2017generic}
\bibfield{author}{\bibinfo{person}{Ishai Rosenberg}, \bibinfo{person}{Asaf
  Shabtai}, \bibinfo{person}{Lior Rokach}, {and} \bibinfo{person}{Yuval
  Elovici}.} \bibinfo{year}{2017}\natexlab{}.
\newblock \showarticletitle{Generic Black-Box End-to-End Attack against RNNs
  and Other API Calls Based Malware Classifiers}.
\newblock \bibinfo{journal}{\emph{arXiv preprint arXiv:1707.05970}}
  (\bibinfo{year}{2017}).
\newblock


\bibitem[\protect\citeauthoryear{Shaham, Yamada, and Negahban}{Shaham
  et~al\mbox{.}}{2015}]%
        {shaham2015understanding}
\bibfield{author}{\bibinfo{person}{Uri Shaham}, \bibinfo{person}{Yutaro
  Yamada}, {and} \bibinfo{person}{Sahand Negahban}.}
  \bibinfo{year}{2015}\natexlab{}.
\newblock \showarticletitle{Understanding Adversarial Training: Increasing
  Local Stability of Neural Nets through Robust Optimization}.
\newblock \bibinfo{journal}{\emph{CoRR}}  \bibinfo{volume}{abs/1511.05432}
  (\bibinfo{year}{2015}).
\newblock
\urldef\tempurl%
\url{http://arxiv.org/abs/1511.05432}
\showURL{%
\tempurl}


\bibitem[\protect\citeauthoryear{Shokri, Stronati, Song, and Shmatikov}{Shokri
  et~al\mbox{.}}{2017}]%
        {shokri2017membership}
\bibfield{author}{\bibinfo{person}{Reza Shokri}, \bibinfo{person}{Marco
  Stronati}, \bibinfo{person}{Congzheng Song}, {and} \bibinfo{person}{Vitaly
  Shmatikov}.} \bibinfo{year}{2017}\natexlab{}.
\newblock \showarticletitle{Membership inference attacks against machine
  learning models}. In \bibinfo{booktitle}{\emph{Security and Privacy (SP),
  2017 IEEE Symposium on}}. IEEE, \bibinfo{pages}{3--18}.
\newblock


\bibitem[\protect\citeauthoryear{Silva}{Silva}{2018}]%
        {medium}
\bibfield{author}{\bibinfo{person}{Thalles Silva}.}
  \bibinfo{year}{2018}\natexlab{}.
\newblock \showarticletitle{An intuitive introduction to Generative Adversarial
  Networks}.
\newblock  (\bibinfo{year}{2018}).
\newblock
\urldef\tempurl%
\url{https://medium.freecodecamp.org/an-intuitive-introduction-to-generative-adversarial-networks-gans-7a2264a81394}
\showURL{%
\tempurl}


\bibitem[\protect\citeauthoryear{Springenberg and Riedmiller}{Springenberg and
  Riedmiller}{2014}]%
        {springenberg2014convnet}
\bibfield{author}{\bibinfo{person}{Dosovitskiy Alexey Brox~Thomas Springenberg,
  Jost~Tobias} {and} \bibinfo{person}{Martin Riedmiller}.}
  \bibinfo{year}{2014}\natexlab{}.
\newblock \showarticletitle{Striving for Simplicity: The All Convolutional
  Net}.
\newblock \bibinfo{journal}{\emph{arXiv preprint arXiv:1412.6806, 2014}}
  (\bibinfo{year}{2014}).
\newblock


\bibitem[\protect\citeauthoryear{Szegedy, Zaremba, Sutskever, Bruna, Erhan,
  Goodfellow, and Fergus}{Szegedy et~al\mbox{.}}{2013}]%
        {szegedy2013intriguing}
\bibfield{author}{\bibinfo{person}{Christian Szegedy},
  \bibinfo{person}{Wojciech Zaremba}, \bibinfo{person}{Ilya Sutskever},
  \bibinfo{person}{Joan Bruna}, \bibinfo{person}{Dumitru Erhan},
  \bibinfo{person}{Ian~J. Goodfellow}, {and} \bibinfo{person}{Rob Fergus}.}
  \bibinfo{year}{2013}\natexlab{}.
\newblock \showarticletitle{Intriguing properties of neural networks}.
\newblock \bibinfo{journal}{\emph{CoRR}}  \bibinfo{volume}{abs/1312.6199}
  (\bibinfo{year}{2013}).
\newblock
\urldef\tempurl%
\url{http://arxiv.org/abs/1312.6199}
\showURL{%
\tempurl}


\bibitem[\protect\citeauthoryear{T.~Ciodaro and Damazio}{T.~Ciodaro and
  Damazio}{2012}]%
        {cio2012structure}
\bibfield{author}{\bibinfo{person}{J.~de~Seixas T.~Ciodaro, D.~Deva} {and}
  \bibinfo{person}{D. Damazio}.} \bibinfo{year}{2012}\natexlab{}.
\newblock \showarticletitle{Online particle detection with neural networks
  based on topological calorimetry information}.
\newblock \bibinfo{journal}{\emph{Journal of physics: conference series}}
  \bibinfo{volume}{238, no. 1} (\bibinfo{year}{2012}).
\newblock


\bibitem[\protect\citeauthoryear{Tram{\`{e}}r, Kurakin, Papernot, Boneh, and
  McDaniel}{Tram{\`{e}}r et~al\mbox{.}}{2017}]%
        {Tramer2017ensemble}
\bibfield{author}{\bibinfo{person}{Florian Tram{\`{e}}r},
  \bibinfo{person}{Alexey Kurakin}, \bibinfo{person}{Nicolas Papernot},
  \bibinfo{person}{Dan Boneh}, {and} \bibinfo{person}{Patrick~D. McDaniel}.}
  \bibinfo{year}{2017}\natexlab{}.
\newblock \showarticletitle{Ensemble Adversarial Training: Attacks and
  Defenses}.
\newblock \bibinfo{journal}{\emph{CoRR}}  \bibinfo{volume}{abs/1705.07204}
  (\bibinfo{year}{2017}).
\newblock
\showeprint[arxiv]{1705.07204}
\urldef\tempurl%
\url{http://arxiv.org/abs/1705.07204}
\showURL{%
\tempurl}


\bibitem[\protect\citeauthoryear{Tram{\`e}r, Zhang, Juels, Reiter, and
  Ristenpart}{Tram{\`e}r et~al\mbox{.}}{2016}]%
        {tramer2016stealing}
\bibfield{author}{\bibinfo{person}{Florian Tram{\`e}r}, \bibinfo{person}{Fan
  Zhang}, \bibinfo{person}{Ari Juels}, \bibinfo{person}{Michael~K Reiter},
  {and} \bibinfo{person}{Thomas Ristenpart}.} \bibinfo{year}{2016}\natexlab{}.
\newblock \showarticletitle{Stealing Machine Learning Models via Prediction
  APIs.}. In \bibinfo{booktitle}{\emph{USENIX Security Symposium}}.
  \bibinfo{pages}{601--618}.
\newblock


\bibitem[\protect\citeauthoryear{Uri~Shaham and Kluger}{Uri~Shaham and
  Kluger}{2018}]%
        {shaham2018basisfunc}
\bibfield{author}{\bibinfo{person}{Yutaro Yamada Ethan Weinberger Alex
  Cloninger Xiuyuan Cheng Kelly~Stanton Uri~Shaham, James~Garritano} {and}
  \bibinfo{person}{Yuval Kluger}.} \bibinfo{year}{2018}\natexlab{}.
\newblock \showarticletitle{Defending against Adversarial Images using Basis
  Functions Transformations}.
\newblock \bibinfo{journal}{\emph{arXiv preprint arXiv:1803.10840}}
  (\bibinfo{year}{2018}).
\newblock


\bibitem[\protect\citeauthoryear{with Adversarial~Examples}{with
  Adversarial~Examples}{2017}]%
        {online:openai}
\bibfield{author}{\bibinfo{person}{OpenAI: Attacking Machine~Learning with
  Adversarial~Examples}.} \bibinfo{year}{2017}\natexlab{}.
\newblock
\newblock


\bibitem[\protect\citeauthoryear{Xu, Evans, and Qi}{Xu et~al\mbox{.}}{2017a}]%
        {xu2017feature}
\bibfield{author}{\bibinfo{person}{Weilin Xu}, \bibinfo{person}{David Evans},
  {and} \bibinfo{person}{Yanjun Qi}.} \bibinfo{year}{2017}\natexlab{a}.
\newblock \showarticletitle{Feature Squeezing: Detecting Adversarial Examples
  in Deep Neural Networks}.
\newblock \bibinfo{journal}{\emph{CoRR}}  \bibinfo{volume}{abs/1704.01155}
  (\bibinfo{year}{2017}).
\newblock
\showeprint[arxiv]{1704.01155}
\urldef\tempurl%
\url{http://arxiv.org/abs/1704.01155}
\showURL{%
\tempurl}


\bibitem[\protect\citeauthoryear{Xu, Evans, and Qi}{Xu et~al\mbox{.}}{2017b}]%
        {xu2017feature2}
\bibfield{author}{\bibinfo{person}{Weilin Xu}, \bibinfo{person}{David Evans},
  {and} \bibinfo{person}{Yanjun Qi}.} \bibinfo{year}{2017}\natexlab{b}.
\newblock \showarticletitle{Feature Squeezing: Detecting Adversarial Examples
  in Deep Neural Networks}.
\newblock \bibinfo{journal}{\emph{CoRR}}  \bibinfo{volume}{abs/1704.01155}
  (\bibinfo{year}{2017}).
\newblock
\showeprint[arxiv]{1704.01155}
\urldef\tempurl%
\url{http://arxiv.org/abs/1704.01155}
\showURL{%
\tempurl}


\bibitem[\protect\citeauthoryear{Xu and Li}{Xu and Li}{2015}]%
        {xu2015leakyrelu}
\bibfield{author}{\bibinfo{person}{Wang Naiyan Chen~Tianqi Xu, Bing} {and}
  \bibinfo{person}{Mu Li}.} \bibinfo{year}{2015}\natexlab{}.
\newblock \showarticletitle{Empirical evaluation of rectified activations in
  convolutional network}.
\newblock \bibinfo{journal}{\emph{arXiv preprint arXiv:1505.00853, 2015}}
  (\bibinfo{year}{2015}).
\newblock


\bibitem[\protect\citeauthoryear{Yan~Zhou}{Yan~Zhou}{2018}]%
        {zhou2018gametheory}
\bibfield{author}{\bibinfo{person}{Bowei~Xi Yan~Zhou, Murat~Kantarcioglu}.}
  \bibinfo{year}{2018}\natexlab{}.
\newblock \showarticletitle{A survey of game theoretic approach for adversarial
  machine learning}.
\newblock \bibinfo{journal}{\emph{Wires Data Mining and Knowledge Discovery}}
  (\bibinfo{year}{2018}).
\newblock


\end{thebibliography}
